# Agentic large language models improve retrieval-based radiology question answering

Sebastian Wind (1,2), Jeta Sopa (1), Daniel Truhn (3), Mahshad Lotfinia (3), Tri-Thien Nguyen (1,4), Keno Bressem (5,6), Lisa Adams (6), Mirabela Rusu (7,8), Harald Köstler (2,9), Gerhard Wellein (2), Andreas Maier (1,2), Soroosh Tayebi Arasteh (1,3,7,8)

(1) Pattern Recognition Lab, Friedrich-Alexander-Universität Erlangen-Nürnberg, Erlangen, Germany.
(2) Erlangen National High Performance Computing Center, Friedrich-Alexander-Universität Erlangen-Nürnberg, Erlangen, Germany.
(3) Department of Diagnostic and Interventional Radiology, University Hospital RWTH Aachen, Aachen, Germany.
(4) Institute of Radiology, University Hospital Erlangen, Erlangen, Germany.
(5) Department of Cardiovascular Radiology and Nuclear Medicine, TUM University Clinic, School of medicine and Health, German Heart Center, Technical University of Munich, Munich, Germany.
(6) Department of Diagnostic and Interventional Radiology, TUM University Clinic, School of Medicine and Health, Klinikum rechts der Isar, Technical University of Munich, Munich, Germany.
(7) Department of Radiology, Stanford University, Stanford, CA, USA.
(8) Department of Urology, Stanford University, Stanford, CA, USA.
(9) Chair of Computer Science 10, Friedrich-Alexander-Universität Erlangen-Nürnberg, Erlangen, Germany.

**Correspondence**

Sebastian Wind, MSc (sebastian.wind@fau.de) or
Soroosh Tayebi Arasteh, PhD, PhD (soroosh.arasteh@rwth-aachen.de)
Pattern Recognition Lab
Friedrich-Alexander-Universität Erlangen-Nürnberg
Martensstr. 3
91058 Erlangen, Germany

# Abstract

Clinical decision-making in radiology increasingly benefits from artificial intelligence (AI), particularly through large language models (LLMs). However, traditional retrieval-augmented generation (RAG) systems for radiology question answering (QA) typically rely on single-step retrieval, limiting their ability to handle complex clinical reasoning tasks. Here we propose radiology Retrieval and Reasoning (RaR), a multi-step retrieval and reasoning framework designed to improve diagnostic accuracy, factual consistency, and clinical reliability of LLMs in radiology question answering. We evaluated 25 LLMs spanning diverse architectures, parameter scales (0.5B to >670B), and training paradigms (general-purpose, reasoning-optimized, clinically fine-tuned), using 104 expert-curated radiology questions from previously established RSNA-RadioQA and ExtendedQA datasets. To assess generalizability, we additionally tested on an unseen internal dataset of 65 real-world radiology board examination questions. RaR significantly improved mean diagnostic accuracy over zero-shot prompting (75% vs. 67%; P = $1.1 \times 10^{-7}$) and conventional online RAG (75% vs. 69%; P = $1.9 \times 10^{-6}$). The greatest gains occurred in mid-sized models (e.g., Mistral Large improved from 72% to 81%) and small-scale models (e.g., Qwen 2.5-7B improved from 55% to 71%), while very large models (>200B parameters) demonstrated minimal changes (<2% improvement). Additionally, RaR reduced hallucinations (mean 9.4%) and retrieved clinically relevant context in 46% of cases, substantially aiding factual grounding. Even clinically fine-tuned models showed gains from RaR (e.g., MedGemma-27B improved from 71% to 81%), indicating that retrieval remains beneficial despite embedded domain knowledge. These results highlight the potential of RaR to enhance factuality and diagnostic accuracy in radiology QA, particularly among mid-sized LLMs, warranting future studies to validate their clinical utility. All datasets, code, and the full RaR framework are publicly available to support open research and clinical translation.

# Introduction

Artificial intelligence (AI) is rapidly transforming diagnostic radiology by enhancing imaging interpretation, improving diagnostic precision, and streamlining clinical workflows[1,2]. Recent advances in large language models (LLMs)[3–7], such as GPT-4[8], have shown remarkable capability in extracting structured information from radiology reports, supporting clinical reasoning, and enabling natural language interfaces[3,9–12]. However, a key limitation persists: the static nature of LLM training data, which may lead to incomplete, outdated, or biased knowledge, thereby compromising clinical accuracy and reliability.

Retrieval-augmented generation (RAG)[13], first introduced by Lewis et al., predates modern large language models and broadly combines generative models with external corpora to ground outputs in retrieved information. When paired with domain-specific knowledge sources, RAG can improve factual accuracy and reduce hallucinations[6,14–17], but its effectiveness depends critically on the quality and coverage of retrieval, and retrieved content is not guaranteed to be correct. Tayebi Arasteh et al. recently introduced Radiology RAG (RadioRAG)[18], an online RAG framework leveraging real-time content from Radiopaedia[19], which demonstrated substantial accuracy improvements in certain LLMs such as GPT-3.5-turbo compared to conventional zero-shot inference. However, these gains were inconsistent, with models like Llama3-8B showing negligible improvements, reflecting limitations in traditional single-step retrieval architectures. Current online RAG frameworks[16,18,20], including RadioRAG[18], primarily employ a single-step retrieval and generation process, limiting their ability to manage complex, multi-part clinical questions[21]. These designs lack iterative refinement, dynamic query expansion, and systematic evaluation of intermediate uncertainty[20]. To address these gaps, multi-step retrieval and reasoning frameworks have recently emerged as an advanced paradigm in AI research[3,22–24]. Recent work in medicine, including i-MedRAG[25], MedAide[26], MedAgentBench[27], and MedChain[28], and more specifically recent works in radiology such as CT-Agent[29] for computed tomography QA, RadCouncil[30] and Yi et al.[31] for report generation, and agent-based uncertainty awareness for report labeling[32] further underscores their growing role in improving factual reliability and interpretability. Such approaches enable LLMs to orchestrate retrieval[33], reasoning, and synthesis in iterative multi-step chains[34,35], supporting dynamic adaptation and enhanced problem-solving capabilities[36–38]. They have shown success across domains such as oncology, general clinical decision-making, and biomedical research[22,23,39], improving both accuracy and interpretability compared to static prompting and conventional RAG. Despite these promising outcomes, their utility in radiology remains largely unexplored, even though radiology uniquely demands nuanced, multi-step reasoning and retrieval of specialized domain knowledge[40].

In this study, we address this crucial gap by systematically evaluating the effectiveness of multi-step retrieval and reasoning in text-based radiology question answering (QA). We introduce RaR, a framework that decomposes clinical questions into structured diagnostic options, retrieves targeted evidence from the comprehensive, peer-reviewed Radiopaedia.org knowledge base, and synthesizes evidence-based responses through iterative reasoning. Using 104 expert-curated radiology questions from the RSNA-RadioQA and ExtendedQA datasets of the RadioRAG study[18] (see **Supplementary Table 1** for dataset characteristics), we compare zero-shot inference, conventional online RAG, and RaR. To assess generalizability, we additionally evaluate RaR on an independent internal dataset of 65 authentic board-style radiology questions from the Technical University of Munich, reflecting real-world assessment conditions and minimizing risk of data leakage. Across 25 diverse LLMs—including proprietary systems (GPT-4-turbo[8], GPT-5, o3), open-weight models (Mistral Large, Qwen 2.5[41]), and clinically fine-tuned variants (MedGemma[42], Llama3-Med42[43])—spanning small (0.5B) to mid-sized (17–110B) and very large architectures (>200B, e.g., DeepSeek-R1[44], o3), we systematically assess the impact of retrieval and reasoning on radiology QA (see **Table 1**). Our results show that RaR consistently enhances diagnostic accuracy and factual reliability across most model classes, with the largest gains in small and mid-sized models where conventional retrieval is insufficient. Very large models (>200B) with strong internal reasoning benefit less, likely due to extensive pretraining and generalization ability, yet even clinically fine-tuned models demonstrate meaningful improvements—suggesting that retrieval and fine-tuning offer complementary strengths. RaR also reduces hallucinations and surfaces clinically relevant content that assists not only LLMs but also radiologists, underscoring its potential to improve factuality, accuracy, and interpretability. **Figure 1** provides an overview of the pipeline, and **Figure 2** illustrates a representative worked example, with additional methodological details in Materials and Methods. Importantly, this study focuses on text-only radiology QA, and future work should extend RaR to multimodal tasks involving imaging data.

## MULTI-STEP LLM RESEARCH PIPELINE ARCHITECTURE

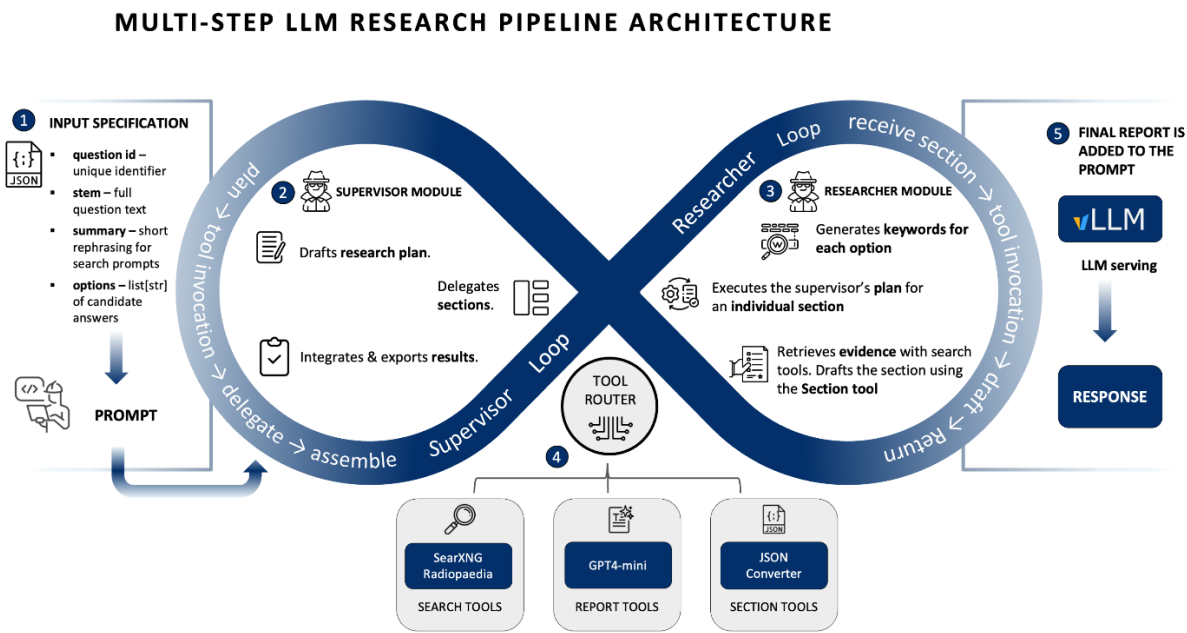

**Figure 1**: **Multi-step architecture of the RaR framework for radiology question answering**. The pipeline combines structured retrieval with multi-step reasoning to generate evidence-grounded diagnostic reports. (1) Each question is preprocessed to extract key diagnostic concepts (using Mistral Large) and paired with multiple-choice options. (2) A supervisor module creates a structured research plan, delegating each diagnostic option to a dedicated research module. (3) Research modules iteratively retrieve targeted evidence from www.radiopaedia.org via a SearXNG-powered search tool, refining queries when needed. (4) Retrieved content is synthesized into structured report sections (using GPT-4o-mini and formatting tools), including supporting and contradicting evidence with citations. (5) The supervisor compiles all sections into a final diagnostic report (introduction, analysis, and conclusion), which is appended to the prompt for final answer selection. The entire workflow is coordinated through a stateful directed graph that preserves shared memory, retrieved context, and intermediate drafts.

**Table 1: Specifications of the language models evaluated in this study.** Summary of the 25 LLMs assessed across zero-shot prompting, conventional online RAG, and the proposed radiology Retrieval and Reasoning (RaR). Listed for each model are parameter count (in billions), training category (e.g., instruction-tuned (IT), reasoning-optimized), accessibility, knowledge cutoff date, developer, and context length (in thousand tokens). Evaluations were conducted between July 1 – August 22, 2025. Note: GPT-5 is included as a widely used system-level benchmark rather than a single fixed model architecture, as it dynamically routes queries across underlying models depending on the task.

| Model name | Parameters (billion) | Category | Accessibility | Knowledge cutoff date | Developer | Context length (thousand tokens) |
|---|---|---|---|---|---|---|
| Ministral-8B | 8 | IT | Open-source | October 2023 | Mistral AI | 128 |
| Mistral Large | 123 | IT | Open-source | November 2024 | Mistral AI | 128 |
| Llama3.3-8B | 8 | IT | Open-weights | March 2023 | Meta AI | 8 |
| Llama3.3-70B | 70 | IT | Open-weights | December 2023 | Meta AI | 128 |
| Llama3-Med42-8B | 8 | IT, clinically-aligned | Open-weights | August 2024 | M42 Health AI Team | 8 |
| Llama3-Med42-70B | 70 | IT, clinically-aligned | Open-weights | August 2024 | M42 Health AI Team | 8 |
| Llama4 Scout 16E | 17 | IT, 17B active parameters | Open-weights | August 2023 | Meta AI | 10,000 (10M) |
| DeepSeek R1-70B | 70 | Reasoning | Open-source | January 2025 | DeepSeek | 128 |
| DeepSeek-R1 | 671 | Reasoning | Open-source | January 2025 | DeepSeek | 128 |
| DeepSeek-V3 | 671 | Mixture of experts | Open-source | July 2024 | DeepSeek | 128 |
| Qwen 2.5-0.5B | 0.5 | IT | Open-source | September 2024 | Alibaba Cloud | 32 |
| Qwen 2.5-3B | 3 | IT | Open-source | September 2024 | Alibaba Cloud | 32 |
| Qwen 2.5-7B | 7 | IT | Open-source | September 2024 | Alibaba Cloud | 131 |
| Qwen 2.5-14B | 14 | IT | Open-source | September 2024 | Alibaba Cloud | 131 |
| Qwen 2.5-70B | 70 | IT | Open-source | September 2024 | Alibaba Cloud | 131 |
| Qwen 3-8B | 8 | Reasoning, mixture of experts | Open-source | December 2024 | Alibaba Cloud | 32 |
| Qwen 3-235B | 235 | Reasoning, mixture of experts | Open-source | July 2025 | Alibaba Cloud | 32 |
| GPT-3.5-turbo | Undisclosed | IT | Proprietary | September 2021 | OpenAI | 16 |
| GPT-4-turbo | Undisclosed | IT | Proprietary | December 2023 | OpenAI | 128 |
| o3 | Undisclosed | Reasoning | Proprietary | June 2024 | OpenAI | 200 |
| GPT-5 | Undisclosed | IT, reasoning | Proprietary | September 2024 | OpenAI | 128 |
| MedGemma-4B-it | 4 | Gemma 3-based, multimodal, IT, clinical reasoning | Open-weights | July 2025 | Google DeepMind | 128 |
| MedGemma-27B-text-it | 27 | Gemma 3-based, text only, IT, clinical reasoning | Open-weights | July 2025 | Google DeepMind | ≥ 128 |
| Gemma-3-4B-it | 4 | IT | Open-weights | August 2024 | Google DeepMind | 128 |
| Gemma-3-27B-it | 27 | IT | Open-weights | August 2024 | Google DeepMind | 128 |

# Results

## Comparison of zero-shot, conventional RAG, and RaR across models

We assessed the diagnostic performance of 25 LLMs across three distinct inference strategies: zero-shot prompting, conventional online RAG, and our proposed RaR framework. The LLMs included: Ministral-8B, Mistral Large, Llama3.3-8B[45,46], Llama3.3-70B[45,46], Llama3-Med42-8B[43], Llama3-Med42-70B[43], Llama4 Scout 16E[33], DeepSeek R1-70B[44], DeepSeek-R1[44], DeepSeek-V3[47], Qwen 2.5-0.5B[41], Qwen 2.5-3B[41], Qwen 2.5-7B[41], Qwen 2.5-14B[41], Qwen 2.5-70B[41], Qwen 3-8B[48], Qwen 3-235B[48], GPT-3.5-turbo, GPT-4-turbo[8], o3, GPT-5[49], MedGemma-4B-it[42], MedGemma-27B-text-it[42], Gemma-3-4B-it[50,51], and Gemma-3-27B-it[50,51]. Accuracy was measured using the 104-question RadioRAG benchmark dataset, with detailed results presented in **Table 2**. When aggregating results across all LLMs, RaR demonstrated a statistically significant improvement in accuracy compared to zero-shot prompting (P = $1.1 \times 10^{-7}$). As previously established, the traditional RAG approach also outperformed zero-shot prompting, showing a smaller but statistically significant gain (P = 0.019). Importantly, RaR further outperformed traditional online RAG (P = $1.9 \times 10^{-6}$), underscoring the benefit of iterative retrieval and autonomous reasoning over single-pass retrieval pipelines. These findings indicate that, at the group level, RaR introduces measurable and additive improvements in radiology question answering, even when compared against established, high-performing RAG systems. The retrieval stage of RaR was guided by a diagnostic abstraction step that condensed each question into key clinical concepts to enable focused evidence search (see **Supplementary Note 1** for examples and implementation details).

## Factual consistency and hallucination rates

To assess factual reliability under RaR, we conducted a hallucination analysis across all 25 LLMs using the 104-question RadioRAG benchmark. Each response was reviewed by a board-certified radiologist (TTN) to evaluate (i) whether the retrieved context was clinically relevant, (ii) whether the model's answer was grounded in that context, and (iii) whether the final output was factually correct. Context was classified as relevant only if it contained no incorrect or off-topic content relative to the diagnostic question, a deliberately strict criterion. Under this definition, clinically relevant evidence was retrieved in 46% of cases (48/104). Detailed results are provided in **Table 3**. To test whether RaR's gains depended on retrieval quality, we repeated the analysis using only the 48 questions with clinically relevant retrieved context. On this subset, RaR significantly improved accuracy across models (68% → 81%; P = $5.1 \times 10^{-9}$; **Supplementary Table 2**), indicating that its benefits persist even under fully correct retrieval. Across the full dataset (104 questions), RaR increased mean accuracy from 67% → 75%. This comparison shows that RaR yields its largest improvements when accurate evidence is available (+13 vs. +8 percentage points overall) while remaining robust to retrieval noise.

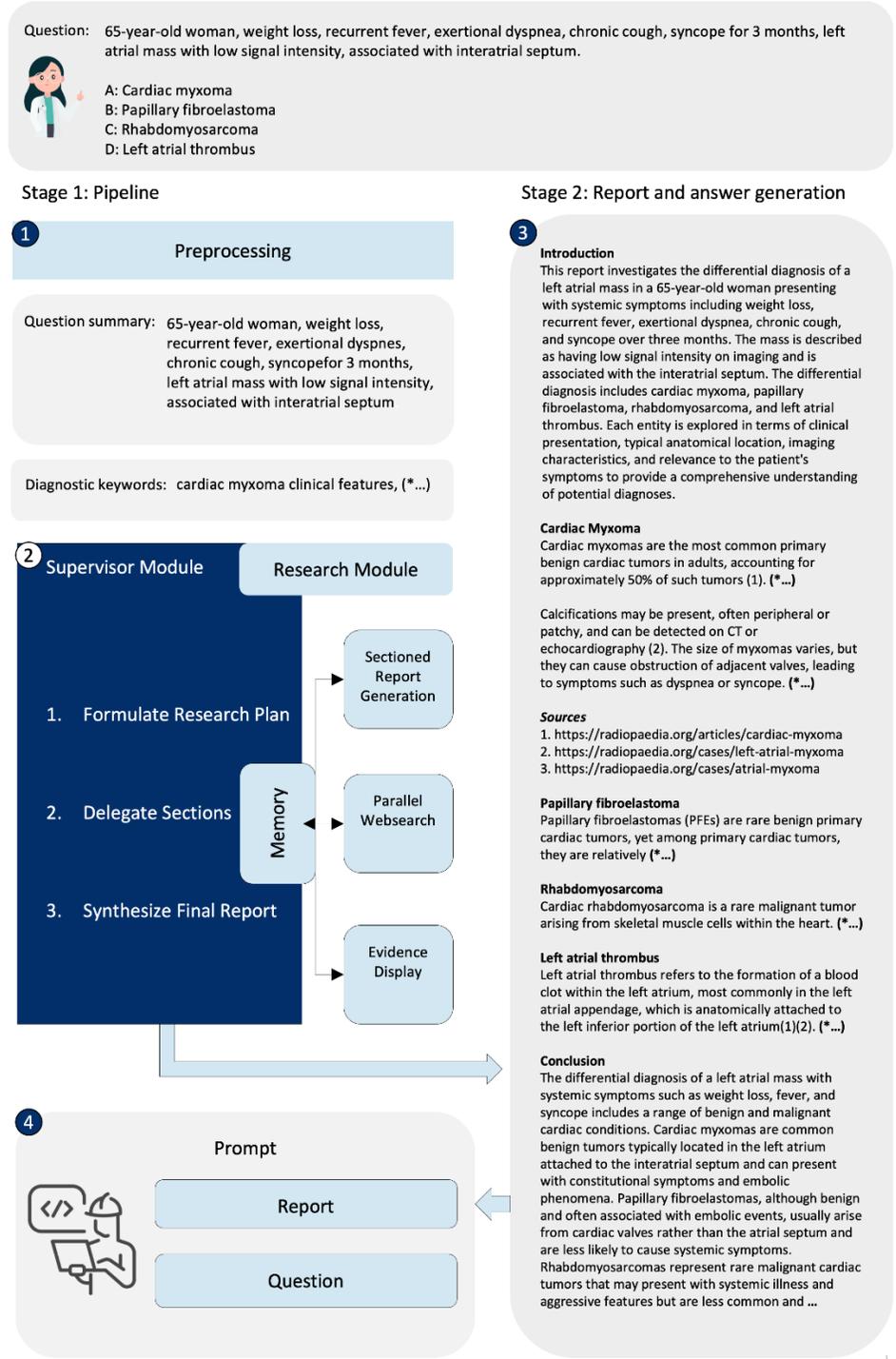

**Figure 2**: **Representative example of the RaR process for a radiology question answering item**. This figure shows the full RaR workflow for a representative question (RSNA-RadioQA-Q53) involving a patient with systemic symptoms and a low signal intensity left atrial mass associated with the interatrial septum. The pipeline begins with keyword-based summarization to guide retrieval, followed by parallel evidence searches for each diagnostic option using Radiopaedia.org. Retrieved content is synthesized into a structured report, including an introduction, citation-backed analyses of all options (cardiac myxoma, papillary fibroelastoma, rhabdomyosarcoma, and left atrial thrombus), and a neutral conclusion. The approach supports interpretable, evidence-grounded radiology question answering.

**Table 2: Accuracy of language models across zero-shot prompting, conventional online RAG, and RaR on the RadioRAG dataset.** Accuracy is reported in percentage as mean ± standard deviation, with 95% confidence intervals shown in brackets. Results are based on 104 questions, using bootstrapping with 1,000 repetitions and replacement while preserving pairing. P-values were calculated for each model using McNemar's test on paired outcomes relative to RaR and adjusted for multiple comparisons using the false discovery rate. A p-value < 0.05 was considered statistically significant. Accuracy is presented alongside total correct answers per method.

| Model name | Zero-shot | | | Conventional online RAG | | | RaR | |
|---|---|---|---|---|---|---|---|---|
| | Accuracy (%) | Total correct (n) | P-value | Accuracy (%) | Total correct (n) | P-value | Accuracy (%) | Total correct (n) |
| Ministral-8B | 47 ± 5 [38, 57] | 49 | 0.020 | 51 ± 5 [41, 61] | 53 | 0.051 | 66 ± 5 [57, 76] | 69 |
| Mistral Large (123B) | 72 ± 4 [63, 81] | 75 | 0.146 | 74 ± 4 [65, 83] | 77 | 0.273 | 81 ± 4 [72, 88] | 84 |
| Llama3.3-8B | 62 ± 5 [53, 71] | 65 | 0.807 | 63 ± 5 [55, 72] | 66 | 0.999 | 65 ± 5 [57, 74] | 68 |
| Llama3.3-70B | 76 ± 4 [67, 84] | 79 | 0.212 | 73 ± 4 [63, 81] | 76 | 0.081 | 83 ± 4 [75, 89] | 86 |
| Llama3-Med42-8B | 67 ± 5 [58, 77] | 70 | 0.263 | 67 ± 5 [59, 77] | 70 | 0.383 | 75 ± 4 [66, 84] | 78 |
| Llama3-Med42-70B | 72 ± 4 [63, 80] | 75 | 0.263 | 75 ± 4 [67, 83] | 78 | 0.705 | 79 ± 4 [71, 87] | 82 |
| Llama4 Scout 16E | 76 ± 4 [67, 85] | 79 | 0.392 | 80 ± 4 [72, 88] | 83 | 0.999 | 81 ± 4 [73, 88] | 84 |
| DeepSeek R1-70B | 78 ± 4 [70, 86] | 81 | 0.859 | 76 ± 4 [67, 84] | 79 | 0.662 | 80 ± 4 [72, 88] | 83 |
| DeepSeek R1 (671B) | 82 ± 4 [74, 89] | 85 | 0.859 | 79 ± 4 [71, 87] | 82 | 0.999 | 80 ± 4 [72, 88] | 83 |
| DeepSeek-V3 (671B) | 76 ± 4 [67, 84] | 79 | 0.106 | 80 ± 4 [72, 88] | 83 | 0.273 | 86 ± 4 [78, 92] | 89 |
| Qwen 2.5-0.5B | 37 ± 5 [27, 46] | 38 | 0.726 | 46 ± 5 [37, 56] | 48 | 0.737 | 42 ± 5 [32, 52] | 43 |
| Qwen 2.5-3B | 54 ± 5 [44, 63] | 56 | 0.146 | 53 ± 5 [43, 62] | 55 | 0.171 | 65 ± 5 [56, 74] | 68 |
| Qwen 2.5-7B | 55 ± 5 [45, 64] | 57 | 0.041 | 59 ± 5 [49, 68] | 61 | 0.171 | 71 ± 4 [62, 80] | 74 |
| Qwen 2.5-14B | 68 ± 4 [59, 77] | 71 | 0.752 | 67 ± 5 [57, 76] | 69 | 0.549 | 72 ± 4 [63, 81] | 75 |
| Qwen 2.5-70B | 70 ± 5 [62, 79] | 73 | 0.185 | 73 ± 4 [64, 82] | 76 | 0.599 | 78 ± 4 [70, 86] | 81 |
| Qwen 3-8B | 66 ± 5 [57, 75] | 69 | 0.157 | 73 ± 4 [65, 81] | 76 | 0.862 | 76 ± 4 [68, 84] | 79 |
| Qwen 3-235B | 82 ± 4 [74, 89] | 85 | 0.999 | 84 ± 4 [75, 90] | 87 | 0.999 | 83 ± 4 [75, 89] | 86 |
| GPT-3.5-turbo | 57 ± 5 [47, 66] | 59 | 0.146 | 62 ± 5 [53, 71] | 64 | 0.540 | 68 ± 5 [60, 77] | 71 |
| GPT-4-turbo | 76 ± 4 [67, 84] | 79 | 0.999 | 76 ± 4 [67, 84] | 79 | 0.999 | 77 ± 4 [69, 85] | 80 |
| o3 | 86 ± 4 [78, 92] | 89 | 0.781 | 85 ± 4 [77, 91] | 88 | 0.705 | 88 ± 3 [81, 93] | 91 |
| GPT-5 | 82 ± 4 [74, 89] | 85 | 0.097 | 80 ± 4 [72, 88] | 83 | 0.081 | 88 ± 3 [82, 94] | 92 |
| MedGemma-4B-it | 56 ± 5 [46, 65] | 58 | 0.157 | 52 ± 5 [42, 62] | 54 | 0.051 | 66 ± 5 [57, 75] | 69 |
| MedGemma-27B-text-it | 71 ± 4 [62, 79] | 74 | 0.146 | 75 ± 4 [66, 84] | 78 | 0.438 | 81 ± 4 [73, 88] | 84 |
| Gemma-3-4B-it | 46 ± 5 [37, 56] | 48 | 0.094 | 53 ± 5 [43, 62] | 55 | 0.273 | 62 ± 5 [52, 71] | 64 |
| Gemma-3-27B-it | 65 ± 5 [57, 75] | 68 | 0.157 | 66 ± 5 [58, 75] | 69 | 0.270 | 76 ± 4 [67, 85] | 79 |

When relevant context was available, most models demonstrated strong factual alignment. Hallucinations, defined as incorrect answers despite the presence of relevant context, occurred in only 9.4% ± 5.9 of questions. The lowest hallucination rates were observed in large-scale and reasoning-optimized models such as o3 (2%), DeepSeek R1 (3%), and GPT-5 (3%), reflecting their superior ability to integrate and interpret retrieved content (see **Figure 3**). In contrast, smaller models such as Qwen 2.5-0.5B (26%) and Gemma-3-4B-it (20%) struggled to do so reliably, exhibiting significantly higher rates of unsupported reasoning.

Interestingly, a substantial proportion of RaR responses were correct despite the retrieved context being clinically irrelevant. On average, 37.4% ± 4.9 of responses fell into this category. This behavior was particularly pronounced among models with strong internal reasoning capabilities, DeepSeek-V3, o3, and Qwen 3-235B each exceeded 40%, suggesting that in the absence of relevant evidence, these models often defaulted to internal knowledge. Similar trends were observed in mid-sized and clinically aligned models, such as Llama3.3-70B, Mistral Large, and MedGemma-27B-text-it, which also maintained high accuracy without external grounding. Conversely, smaller models like Qwen 2.5-0.5B (21%) and Ministral-8B (35%) were less effective under these conditions, indicating greater dependence on successful retrieval.

Across models, an average of 14.3% ± 6.5 of questions were answered incorrectly under zero-shot prompting but correctly after RaR, highlighting the additive diagnostic value of structured evidence acquisition. **Supplementary Tables 3** and **4** provide example responses from GPT-3.5-turbo with and without RaR, alongside the corresponding retrieved content. These findings indicate that RaR improves factual grounding and reduces hallucination by enabling structured, clinically aware evidence refinement. However, model behavior in the absence of relevant context varies substantially, with larger and reasoning-tuned models demonstrating greater resilience through fallback internal reasoning. Representative examples of such cases, including model outputs that were correct despite irrelevant or noisy retrieval, are provided in **Supplementary Note 2**.

To better understand the sources of model errors, we performed a qualitative error analysis across representative cases (see **Supplementary Note 3**). Three dominant error types were identified: reasoning shortcut errors, where models relied on familiar diagnostic patterns instead of verifying the retrieved evidence; context integration errors, where models correctly interpreted individual findings but failed to synthesize them into a coherent diagnosis; and context independence errors, where models produced correct answers despite disregarding the evidence. Overall, RaR markedly reduced shortcut and integration errors by promoting explicit evidence verification and contextual reasoning, correcting approximately 14.3% of previously wrong zero-shot answers.

**Table 3: Hallucination and relevance metrics for RaR-powered responses on the RadioRAG dataset (n = 104).** "Context relevant" was evaluated at the dataset level: each question was labeled as having relevant or irrelevant retrieved context, and the same label was applied across all models (48/104 questions were judged to have clinically appropriate context). "Hallucination" refers to incorrect model answers despite relevant context. "Correct despite irrelevant context" captures correct answers when the retrieved context was not clinically useful. The final column reports the percentage of questions that were incorrect in zero-shot prompting but answered correctly using RaR.

| Model name | Context relevant | Hallucination (relevant context, incorrect response) | Correct despite irrelevant context | Zero-shot incorrect → RaR correct |
|---|---|---|---|---|
| Ministral-8B | 46% (48/104) | 14% (15/104) | 35% (36/104) | 26% (27/104) |
| Mistral Large (123B) | 46% (48/104) | 6% (6/104) | 40% (42/104) | 12% (13/104) |
| Llama3.3-8B | 46% (48/104) | 17% (18/104) | 37% (38/104) | 12% (13/104) |
| Llama3.3-70B | 46% (48/104) | 6% (6/104) | 42% (44/104) | 11% (11/104) |
| Llama3-Med42-8B | 46% (48/104) | 11% (11/104) | 39% (41/104) | 16% (17/104) |
| Llama3-Med42-70B | 46% (48/104) | 7% (7/104) | 39% (41/104) | 12% (13/104) |
| Llama4 Scout 16E | 46% (48/104) | 5% (5/104) | 39% (41/104) | 9% (9/104) |
| DeepSeek R1-70B | 46% (48/104) | 5% (5/104) | 38% (40/104) | 8% (8/104) |
| DeepSeek R1 (671B) | 46% (48/104) | 3% (3/104) | 37% (38/104) | 6% (6/104) |
| DeepSeek-V3 (671B) | 46% (48/104) | 4% (4/104) | 43% (45/104) | 12% (13/104) |
| Qwen 2.5-0.5B | 46% (48/104) | 26% (27/104) | 21% (22/104) | 21% (22/104) |
| Qwen 2.5-3B | 46% (48/104) | 13% (14/104) | 33% (34/104) | 21% (22/104) |
| Qwen 2.5-7B | 46% (48/104) | 12% (12/104) | 37% (38/104) | 23% (24/104) |
| Qwen 2.5-14B | 46% (48/104) | 10% (10/104) | 36% (37/104) | 15% (16/104) |
| Qwen 2.5-70B | 46% (48/104) | 5% (5/104) | 37% (38/104) | 12% (13/104) |
| Qwen 3-8B | 46% (48/104) | 6% (6/104) | 36% (37/104) | 17% (18/104) |
| Qwen 3-235B | 46% (48/104) | 5% (5/104) | 41% (43/104) | 6% (6/104) |
| GPT-3.5-turbo | 46% (48/104) | 13% (14/104) | 36% (37/104) | 21% (22/104) |
| GPT-4-turbo | 46% (48/104) | 9% (9/104) | 39% (41/104) | 8% (8/104) |
| o3 | 46% (48/104) | 2% (2/104) | 43% (45/104) | 3% (3/104) |
| GPT-5 | 46% (48/104) | 3% (3/104) | 45% (47/104) | 7% (7/104) |
| MedGemma-4B-it | 46% (48/104) | 17% (18/104) | 38% (39/104) | 20% (21/104) |
| MedGemma-27B-text-it | 46% (48/104) | 3% (3/104) | 38% (39/104) | 15% (16/104) |
| Gemma-3-4B-it | 46% (48/104) | 20% (21/104) | 36% (37/104) | 25% (26/104) |
| Gemma-3-27B-it | 46% (48/104) | 7% (7/104) | 37% (38/104) | 20% (21/104) |
| *Average* | *46% ± 0* | *9.2% ± 6.1%* | *37.4% ± 4.9%* | *14.3% ± 6.5%* |

# Retrieval performance stratified by model scale: small-scale models

We next assessed whether model size influences the effectiveness of RaR in radiology question answering (see **Figure 4**). Across the seven smallest models in our study (including Ministral-8B, Gemma-3-4B-it, Qwen 2.5-7B, Qwen 2.5-3B, Qwen 2.5-0.5B, Qwen 3-8B, and Llama-3-8B), we observed a consistent trend: conventional online RAG outperformed zero-shot prompting (P = 0.002), and RaR further improved over both baselines (P = 0.016 vs. zero-shot; P = 0.035 vs. traditional online RAG). When examining individual models, only two of the seven demonstrated statistically significant improvements with RaR compared to zero-shot prompting: Qwen 2.5-7B (71% ± 4 [95% CI: 62, 80] vs. 55% ± 5 [95% CI: 45, 64]; P = 0.041) and Ministral-8B (66% ± 5 [95% CI: 57, 76] vs. 47% ± 5 [95% CI: 38, 57]; P = 0.020). The remaining models exhibited absolute accuracy improvements ranging from 3% to 16%, though these did not reach statistical significance after correction for multiple comparisons.

These findings suggest that RaR can enhance performance in small-scale LLMs. However, the degree of benefit varied across models, likely reflecting differences in pretraining data, instruction tuning, and architectural design, even within a similar parameter range.

# Retrieval performance stratified by model scale: large-scale models

We next evaluated the effect of RaR on the largest LLMs in our study, comprising DeepSeek-R1, DeepSeek-V3, o3, Qwen 3-235B, GPT-4-turbo, and GPT-5, all likely to be exceeding 200 billion parameters. These models demonstrated strong performance under zero-shot prompting alone, achieving diagnostic accuracies ranging from 76% to 86% on the RadioRAG benchmark (**Table 2**). Neither conventional online RAG (P = 0.999) nor RaR (P = 0.147) led to meaningful improvements.

Across all five models, accuracy differences between the three inference strategies were minimal (see **Figure 4**). For example, DeepSeek-R1 performed at 82% ± 4 [95% CI: 74, 89] with zero-shot, 80% ± 4 [95% CI: 72, 88] with RaR, and 79% ± 4 [95% CI: 71, 87] with conventional online RAG; o3 improved marginally from 86% ± 4 [95% CI: 78, 92] to 88% ± 3 [95% CI: 81, 93] with RaR; and Qwen3-235B and GPT-4-turbo showed ≤1% changes across conditions. DeepSeek-V3 and GPT-5 showed slightly higher improvement (DeepSeek-V3: from 76% ± 4 [95% CI: 67, 84] to 86% ± 4 [95% CI: 78, 92]; GPT-5: from 82% ± 4 [95% CI: 74, 89] to 88% ± 3 [95% CI: 82, 94], respectively) but still not significant. Traditional RAG showed similarly negligible differences.

These findings indicate that very large LLMs can already handle complex radiology QA tasks with high accuracy without requiring external retrieval. This likely reflects their extensive pretraining on large-scale corpora, improved reasoning abilities, and domain-general coverage, diminishing the marginal value of either conventional RAG or RaR in high-performing settings.

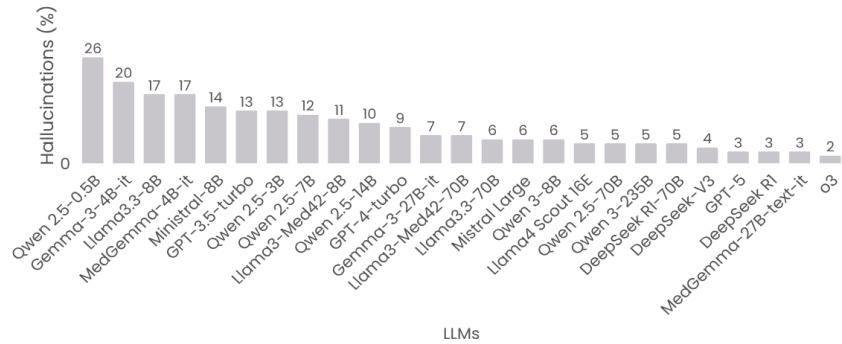

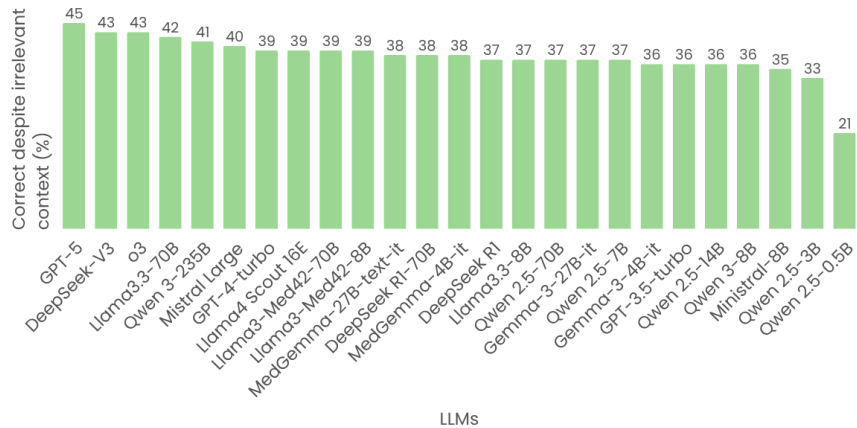

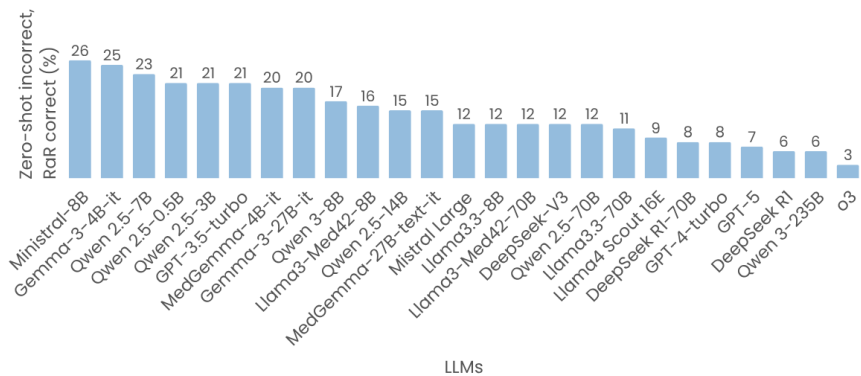

**Figure 3: Factuality assessment of LLM responses on the RadioRAG dataset.** Each bar plot shows the proportion of cases per model falling into a specific factuality category, with models ordered by descending percentage. Comparisons were based on the RadioRAG benchmark dataset (n = 104). **(a)** Hallucinations: Cases in which the provided context was relevant, but the model still generated an incorrect response (context = 1, response = 0). **(b)** Context irrelevance tolerance: Cases where the model produced a correct response despite the retrieved context being unhelpful or irrelevant (context = 0, response = 1). **(c)** RaR correction: Instances where the zero-shot response was incorrect but RaR strategy successfully produced a correct response (zero-shot = 0, RaR = 1).

# Retrieval performance stratified by model scale: mid-sized models

Mid-sized models, typically ranging between 17B and 110B parameters, represent a particularly relevant category for clinical deployment, offering a favorable trade-off between performance and computational efficiency. This group in our study included GPT-3.5-turbo, Llama 3.3-70B, Mistral Large, Qwen 2.5-70B, Llama 4 Scout 16E, Gemma-3-27B-it, and DeepSeek-R1-70B. Across this cohort, the conventional online RAG framework did not yield a statistically significant improvement in accuracy over zero-shot prompting (P = 0.253). In contrast, RaR significantly outperformed both zero-shot (P = 0.001) and conventional online RAG (P = 0.002), suggesting that the benefits of RaR become more apparent in this model size range, where LLMs are strong enough to follow reasoning chains but may still benefit from structured multi-step guidance. While every model in this group showed an absolute improvement in diagnostic accuracy with RaR, for example, GPT-3.5-turbo improved from 57% to 68%, Llama 3.3-70B from 76% ± 4 [95% CI: 67, 84] to 83% ± 4 [95% CI: 75, 89], and Mistral Large from 72% ± 4 [95% CI: 63, 81] to 81% ± 4 [95% CI: 73, 88], none of these increases reached statistical significance when evaluated individually (see **Figure 4**).

To further probe the relationship between model scale and accuracy, we conducted a targeted scaling experiment using the Qwen 2.5 model family, which spans a wide range of sizes (Qwen 2.5-70B, 14B, 7B, 3B, and 0.5B) while maintaining consistent architecture and training procedures. This allowed us to isolate the influence of model size from confounding variables such as instruction tuning or pretraining corpus. We computed Pearson correlation coefficients between model size and diagnostic accuracy for each inference strategy. All three methods including zero-shot (r = 0.68), conventional online RAG (r = 0.81), and RaR (r = 0.61) showed strong positive correlations with parameter count, reflecting the general performance advantage of larger models. However, as detailed in earlier findings, the relative benefit of retrieval strategies was not uniformly distributed: conventional RAG was most beneficial for small models, while RaR consistently enhanced performance in mid-sized models (see **Figure 4**). These findings highlight the importance of aligning retrieval strategies with model capacity and deployment constraints.

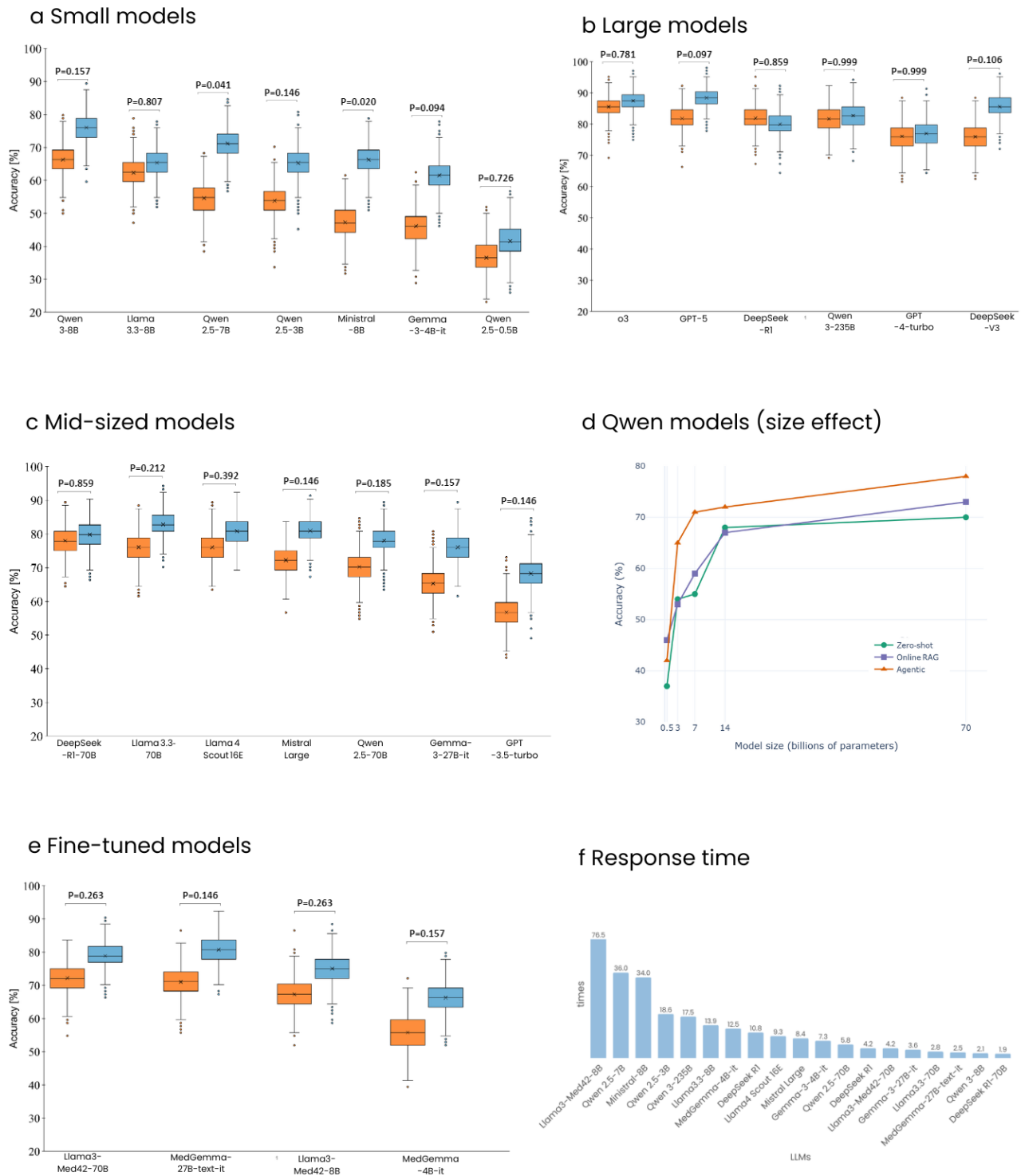

**Figure 4: Comparative accuracy distributions and inference-time multipliers for zero-shot versus RaR strategies across model groups (RadioRAG dataset)**. Accuracy results are shown for **(a)** small-scale models, **(b)** large, **(c)** mid-sized models, **(d)** across Qwen 2.5 family for different parameter sizes: Qwen 2.5-70B, 14B, 7B, 3B and 0.5B, and **(e)** medically fine-tuned models. **(f)** Distribution of RaR-to-zero-shot runtime multipliers (× slower/faster) across all models. comparisons were performed on the RadioRAG benchmark dataset (n = 104). Line chart shows mean accuracy versus model size for zero-shot (green), online RAG (orange) and RaR (purple) across Qwen 2.5 family. P-values were calculated between each pair's accuracy values for each model.

# Effect of clinical fine-tuning on retrieval-augmented performance

To examine whether domain-specific fine-tuning diminishes the utility of retrieval-based strategies, we evaluated four clinically optimized language models: MedGemma-27B-text-it, MedGemma-4B-it, Llama3-Med42-70B, and Llama3-Med42-8B. These models are specifically fine-tuned for biomedical or radiological applications, making them suitable test cases for understanding the complementary role of retrieval and reasoning. Despite already possessing clinical specialization, all four models exhibited improved diagnostic QA performance under RaR. On average, accuracy increased from 67% ± 6 under zero-shot prompting to 75% ± 6 with RaR (P = 0.001). Conventional online RAG, in contrast, did not show a significant improvement over zero-shot prompting (67% ± 9 vs. 67% ± 6, P = 0.704). Notably, RaR also significantly outperformed conventional online RAG (P = 0.034), suggesting that structured multi-step reasoning contributes meaningfully even when baseline knowledge is embedded through fine-tuning. Each model in this group followed a similar pattern. For instance, MedGemma-27B-text-it improved from 71% ± 4 [95% CI: 62, 79] to 81% ± 4 [95% CI: 73, 88] with RaR, MedGemma-4B-it from 56% ± 5 [95% CI: 46, 65] to 66% ± 5 [95% CI: 57, 75], Llama3-Med42-70B from 72% ± 4 [95% CI: 63, 80] to 79% ± 4 [95% CI: 71, 87], and Llama3-Med42-8B from 67% ± 5 [95% CI: 58, 77] to 75% ± 4 [95% CI: 66, 84] (see **Figure 4**). While these individual gains were not statistically significant on their own, the collective improvement supports the hypothesis that retrieval-augmented reasoning provides additive benefits beyond those conferred by fine-tuning alone.

# Latency and computational overhead

To evaluate the computational impact of RaR, we measured and compared per-question response times between zero-shot prompting and RaR across all models using the RadioRAG benchmark. As shown in **Table 4**, RaR introduced a substantial latency overhead across all model groups, with the average response time increasing from 54 ± 28 seconds under zero-shot prompting to 324 ± 270 seconds under RaR, equivalent to a 6.71× increase.

As shown in **Figure 4**, this increase varied considerably by model group. Small-scale models (7–8B parameters), including Qwen 2.5-7B, Qwen3-8B, Llama3-Med42-8B, Llama3-Med42-8B, and Ministral-8B, showed a 6.04× average increase, with individual models ranging from modest (2.06× for Qwen3-8B) to substantial (35.98× for Qwen 2.5-7B). Mini models (3–4B parameters), such as Gemma-3-4B-it, MedGemma-4B-it, and Qwen 2.5-3B, exhibited the highest relative increase, averaging 11.10×, with Qwen2.5-3B peaking at 18.59×. In contrast, mid-sized models (~70B parameters), including DeepSeek-R1-70B, Llama-3.3-70B, Qwen 2.5-70B, and Llama3-Med42-70B, had a more moderate increase of 2.93×. This reflects a balance between computational capacity and the overhead introduced by iterative reasoning. For example, DeepSeek-R1-70B showed only a 1.87× increase. The large-model group (120–250B), including Qwen 3-235B, Mistral Large, and Llama4 Scout 16E, had the largest absolute latency, with a group average increase of 13.27×. Qwen3-235B showed the most pronounced jump, from 97 seconds to 1703 seconds per question. Despite high computational costs, these models showed only minimal diagnostic improvement with RaR, emphasizing a potential efficiency–performance trade-off. Notably, the DeepSeek mixture of experts[52] (MoE) group (DeepSeek-R1 and DeepSeek-V3) exhibited relatively efficient scaling under RaR, with an average increase of 4.19×, suggesting that sparsely activated architectures may offer runtime advantages in multi-step retrieval tasks. Similarly, the Gemma-27B group (Gemma-3-27B-it and MedGemma-27B-text-it) demonstrated a low variance and consistent response time increase of 2.82×, indicating reliable timing behavior under RaR workflow.

Despite these increases, the absolute response times remained within feasible limits for many clinical applications. Furthermore, because evaluations were conducted under identical system conditions, the relative timing metrics provide a robust measure of computational scaling. These findings suggest that while the RaR introduces additional latency, its time cost may be acceptable, especially in mid-sized and sparse-activation models depending on deployment requirements and accuracy demands.

## Effect of retrieved context on human diagnostic accuracy

To better understand the source of diagnostic improvements conferred by RaR, we conducted an additional experiment involving a board-certified radiologist (TTN) with seven years of experience in diagnostic and interventional radiology. As in previous evaluations, the expert first answered all 104 RadioRAG questions unaided, i.e., without access to external references or retrieval assistance, achieving an accuracy of 51% ± 5 [95% CI: 41, 62] (53/104). This baseline performance was significantly lower than that of 17 out of 25 evaluated LLMs in their zero-shot mode ($P ≤ 0.017$), and not significantly different from 7 models, including GPT-3.5-turbo, Llama3.3-8B, Qwen 2.5-7B, Ministral-8B, MedGemma-4B-it, Gemma-3-4B-it, and Qwen 2.5-3B. Only Qwen 2.5-0.5B, the smallest model tested, performed significantly inferior to the radiologist (37% ± 5 [95% CI: 27, 46]; $P = 0.008$).

To isolate the contribution of retrieval independent of generative reasoning, we repeated the experiment with the same radiologist using the contextual reports retrieved by RaR, that is, the same Radiopaedia content supplied to the LLMs. With access to this structured evidence, the radiologist's accuracy increased to 68% ± 5 [95% CI: 60, 77] (71/104), a significant improvement over the unaided baseline ($P = 0.010$). This finding demonstrates that RaR successfully retrieves clinically meaningful and decision-relevant information, which can support human diagnostic accuracy even in the absence of language model synthesis.

When comparing the radiologist's context-assisted performance to that of the LLMs, only 1 out of 25 models significantly outperformed the radiologist under zero-shot conditions (o3; $P = 0.018$). In contrast, when compared to LLM performance under the full RaR framework, only 3 models, i.e., GPT-5 ($P = 0.008$), DeepSeek-V3 ($P = 0.012$) and o3 ($P = 0.008$) achieved statistically significant improvements over the context-assisted radiologist.

**Table 4: Response time comparison between zero-shot and RaR strategies on the RadioRAG dataset.** Average per-question response times (n = 104) are reported in seconds as mean ± standard deviation for both individual models and aggregated model groups. On the RadioRAG dataset, a fixed overhead of 10,554.6 seconds per model, corresponding to context generation, was evenly distributed across all questions, contributing approximately 101.5 seconds per question. For time analysis, models were grouped based on parameter scale and architectural characteristics into six categories: the DeepSeek mixture of experts (MoE) group, the large model group (120–250B), the medium-scale group (~70B), the Gemma-27B group, the small model group (7–8B), and the mini model group (3–4B). "Absolute difference" denotes the increase in average response time per question introduced by the RaR method, and "Relative increase" refers to the ratio of mean RaR time to mean zero-shot time per group. Final statistics are computed at the group level.

| Model / group name | Time | | | |
|---|---|---|---|---|
| | Zero-shot (s) | RaR (s) | Absolute difference (s) | Relative increase (times) |
| **DeepSeek-V3 group** | **98.55 ± 53.58** | **412.7 ± 156.7** | **314.2 ± 141.6** | **4.2 x** |
| **Large (120 – 250B) group** | **63.7 ± 29.4** | **845.1 ± 744.7** | **781.4 ± 715.2** | **13.3 x** |
| Llama4 Scout 16E | 49.6 ± 24.6 | 462.3 ± 190.2 | 412.6 ± 169.7 | 9.3 x |
| Mistral Large | 43.9 ± 23.9 | 369.7 ± 142.0 | 325.8 ± 126.0 | 8.4 x |
| Qwen 3-235B | 97.5 ± 54.6 | 1703.3 ± 787.6 | 1605.8 ± 744.0 | 17.5 x |
| **Medium (≈ 70B) group** | **78.7 ± 51.4** | **230.58 ± 44.8** | **151.8 ± 34.3** | **2.9 x** |
| DeepSeek R1-70B | 151.3 ± 83.4 | 282.8 ± 95.0 | 131.3 ± 68.3 | 1.9 x |
| Llama3-Med42-70B | 42.2 ± 22.4 | 177.0 ± 39.5 | 134.8 ± 27.9 | 4.2 x |
| Llama3.3-70B | 78.5 ± 43.6 | 216.7 ± 60.7 | 138.2 ± 34.7 | 2.8 x |
| Qwen 2.5-70B | 42.6 ± 22.2 | 245.7 ± 76.8 | 203.1 ± 58.5 | 5.8 x |
| **Gemma 27B group** | **75.8 ± 38.2** | **214.1 ± 54.9** | **138.3 ± 16.7** | **2.8 x** |
| Gemma-3-27B-it | 48.8 ± 28.6 | 175.3 ± 37.4 | 126.5 ± 26.2 | 3.6 x |
| MedGemma-27B-text-it | 102.8 ± 56.1 | 253.0 ± 75.2 | 150.1 ± 38.4 | 2.5 x |
| **Small (7 – 8B) group** | **22.0 ± 39.9** | **132.9 ± 33.9** | **110.9 ± 9.3** | **6.0 x** |
| Llama3-Med42-8B | 1.4 ± 0.7 | 108.0 ± 3.7 | 106.6 ± 3.3 | 76.5 x |
| Llama3.3-8B | 8.4 ± 4.0 | 116.3 ± 7.6 | 107.9 ± 4.6 | 13.9 x |
| Ministral-8B | 3.7 ± 2.2 | 124.9 ± 11.8 | 121.2 ± 10.4 | 34.0 x |
| Qwen 2.5-7B | 3.4 ± 1.6 | 122.8 ± 11.4 | 119.4 ± 10.4 | 36.0 x |
| Qwen 3-8B | 93.2 ± 53.4 | 192.3 ± 49.8 | 99.1 ± 33.9 | 2.1 x |
| **Mini (3 – 4B) group** | **11.4 ± 5.4** | **126.3 ± 6.3** | **114.9 ± 8.4** | **11.1 x** |
| Gemma-3-4B-it | 17.5 ± 7.9 | 127.7 ± 13.1 | 110.2 ± 7.0 | 7.3 x |
| MedGemma-4B-it | 9.6 ± 5.4 | 119.4 ± 9.9 | 109.8 ± 9.1 | 12.5 x |
| Qwen 2.5-3B | 7.1 ± 3.7 | 131.7 ± 13.7 | 124.6 ± 11.0 | 18.6 x |
| *Average* | *53.7 ± 28.4* | *324.4 ± 270.2* | *271.2 ± 257.3* | *6.7 ± 4.1 x* |

# Generalization on an independent dataset

To assess generalizability beyond the RadioRAG benchmark, we evaluated all 25 LLMs on an independent internal dataset comprising 65 authentic radiology board examination questions from the Technical University of Munich. These questions were not included in model training or prompting and reflect real-world clinical exam conditions. Results are shown in **Supplementary Figure 1**. RaR again outperformed zero-shot prompting, with average accuracy increasing from 81% ± 14 to 88% ± 8 (P = 0.002). This replicates the overall trend observed in the main benchmark. The gain was statistically significant in small models (P = 0.010), but not in mid-sized (P = 0.174), fine-tuned (P = 0.238), or large models (P = 0.953), a contrast to the benchmark where mid-sized and fine-tuned models also showed significant improvements. This discrepancy may reflect reduced statistical power due to the smaller sample size or differences in question distribution (see **Supplementary Note 4** for subgroup precision and effect size analysis).

To assess factual reliability, we replicated our hallucination analysis on the internal dataset using the same annotation protocol as in the RadioRAG benchmark. Clinically relevant evidence was retrieved in 74% (48/65) of cases, a substantial increase from the 46% observed in the main dataset. This likely reflects the more canonical phrasing and structured nature of board-style questions, which facilitate more effective document matching. Despite the higher relevance rate, hallucination rates remained consistent: the average hallucination rate, defined as incorrect answers despite clinically relevant context, was 9.2% ± 5.5%, nearly identical to the 9.2% ± 6.1 observed in the RadioRAG benchmark. Larger and reasoning-optimized models such as GPT-4-turbo (9%), DeepSeek R1 (8%), and o3 (9%) maintained their strong factual grounding, while smaller models continued to struggle, for example, Qwen 2.5-0.5B hallucinated in 32% of cases even when provided with relevant context. These results confirm that the factual consistency of RaR generalizes well across datasets, with stable hallucination behavior observed across model families. Full model-level hallucination metrics are provided in **Supplementary Table 5**.

To evaluate computational overhead, we repeated the time analysis on the internal dataset (n = 65). On the internal dataset, as shown in **Supplementary Table 6**, RaR inference increased average per-question response time from 35.0 ± 22.9 seconds under zero-shot prompting to 167.5 ± 59.4 seconds under RaR, an absolute increase of 132.4 ± 41.7 seconds, corresponding to a 6.9× ± 4.2 slowdown. These results are consistent with the RadioRAG dataset, which showed a comparable 6.7× ± 4.1 increase. Despite the smaller question set, relative latency patterns across model families remained stable: mini models (3–4B) showed the highest increase (13.7×), followed by small models (10.2×) and large models (5.9×), while mid-sized (~70B) and Gemma-27B groups demonstrated more efficient scaling (4.5× and 3.0×, respectively). The DeepSeek MoE group also maintained efficient performance (3.9×).

To benchmark human diagnostic performance on the internal dataset, we evaluated the same board-certified radiologist (TTN) under two conditions: zero-shot answering and context-assisted answering using only the retrieved evidence from the RaR system. The radiologist achieved 74% ± 5 accuracy under zero-shot conditions, which increased to 85% ± 4 when supported by retrieved context, although this improvement did not reach statistical significance (P

= 0.065). This contrasts with the main RadioRAG dataset, where context significantly boosted the radiologist's accuracy (P = 0.010). The diminished statistical effect in the internal dataset is likely attributable to both the higher baseline accuracy and the smaller sample size (n = 65), reducing the measurable headroom and statistical power, respectively. When compared directly to LLM performance, 7 out of 25 models significantly outperformed the radiologist under zero-shot prompting (P ≤ 0.014), fewer than in the RadioRAG dataset (17/25). However, when both the human and the models were given access to the same retrieved context, no model significantly outperformed the radiologist (P ≥ 0.487), replicating the trend observed in the main dataset (3/25).

# Discussion

In this study, we introduced RaR, a radiology-specific retrieval and reasoning framework designed to enhance the performance, factual grounding, and clinical reliability of LLMs in radiology QA tasks. To the best of our knowledge, our large-scale evaluation across 25 diverse LLMs, including different architectures, parameter scales, training paradigms, and clinical fine-tuning, represents one of the most comprehensive comparative analysis of its kind to date[53]. Our findings indicate that RaR can improve diagnostic accuracy relative to conventional zero-shot prompting and conventional RAG approaches, especially in small- to mid-sized models, while also reducing hallucinated outputs. However, the benefits of RaR were not uniformly observed across all models or scenarios, underscoring the need for careful consideration of model scale and characteristics when deploying retrieval-based systems.

A central finding of this study is that the effectiveness of retrieval strategies strongly depends on model scale. While traditional single-step online RAG[16,18,20], and generally non-agentic RAG[16,17,54,55], approaches have previously been shown to primarily benefit smaller models (<8 billion parameters) with diminishing returns at larger scales[16,18,20], our RaR framework expanded performance improvements into the mid-sized model range (approximately 17–150 billion parameters). Mid-sized models such as GPT-3.5-turbo, Mistral Large, and Llama3.3-70B have sufficient reasoning capabilities to follow structured logic but frequently struggle to independently identify and incorporate relevant external clinical evidence. By decomposing complex clinical questions into structured subtasks and iteratively retrieving targeted evidence, RaR consistently improved accuracy across these mid-sized models, gains that conventional RAG did not achieve in this important segment. Similarly, smaller models also benefited from structured retrieval, overcoming some limitations associated with fewer parameters and less comprehensive pretraining. However, the magnitude of improvements varied between individual small-scale models, likely reflecting differences in architectural design, instruction tuning, and pretraining data. These results suggest that while RaR can broadly enhance performance across smaller and mid-sized models, model-specific optimizations may be required to fully capitalize on its potential.

In contrast, the largest evaluated models (more than 200 billion parameters), such as GPT-5, o3, DeepSeek-R1, and Qwen-3-235B, exhibited minimal to no gains from either conventional or RaR methods. These models achieved high performance with zero-shot inference alone, suggesting that their extensive pretraining on large-scale and potentially clinically relevant data already equipped them with substantial internal knowledge. Beyond pretraining coverage, additional factors likely contribute to this saturation effect. Very large models are known to possess advanced reasoning capabilities, robust in-context learning, and architectural enhancements such as deeper transformer stacks or mixture-of-experts routing, which collectively reduce reliance on external retrieval. These mechanisms may allow large models to internally simulate multi-step reasoning without explicit retrieval augmentation. While retrieval therefore offered limited incremental accuracy benefits at this scale, it may still provide value in clinical practice by enhancing transparency, auditability, and alignment with established documentation standards. Future studies should explore whether RaR can improve interpretability, consistency, and traceability of decisions made by these high-capacity models, even when raw accuracy does not substantially increase.

To further examine the relationship between model scale and retrieval benefit, we conducted a controlled scaling analysis using the Qwen 2.5 model family. This approach, which held architecture and training constant, revealed a strong positive relationship between model size and diagnostic accuracy across all tested inference strategies[56,57]. Nevertheless, the optimal retrieval approach varied: traditional single-step RAG offered the greatest advantage for smaller models, whereas RaR consistently enhanced mid-sized model performance. These results highlight the importance of aligning retrieval strategies with the intrinsic reasoning capacity of individual models, emphasizing tailored rather than universal implementation of retrieval augmentation.

A key consideration in clinical applications is whether domain-specific fine-tuning reduces the necessity or utility of external retrieval. Clinically specialized LLMs, such as variants of MedGemma and Llama3-Med42, are often assumed to contain embedded medical knowledge sufficient for diagnostic reasoning[6]. However, our results show that even these fine-tuned models consistently benefited from RaR: across all four tested models, performance significantly improved when structured evidence was introduced. Nevertheless, fine-tuning itself did not consistently improve diagnostic accuracy compared to general-domain counterparts of similar scale. For example, Llama3-Med42-70B underperformed relative to the non-specialized Llama3.3-70B, despite its radiology-specific adaptation. This finding lends support to concerns that fine-tuning, especially when not carefully balanced, may introduce trade-offs such as catastrophic forgetting or reduced general reasoning ability. Taken together, our results suggest that RaR remains essential even in specialized models, and that domain-specific fine-tuning should not be assumed to universally enhance performance. Instead, retrieval and fine-tuning may offer partially complementary benefits, but their interaction appears model- and implementation-dependent, warranting further empirical scrutiny.

These findings also carry practical implications for model selection. For institutions with limited computational resources, RaR enables smaller and mid-sized models to achieve diagnostic accuracy closer to that of much larger systems, making them a cost-effective option. Very large models (>200B) deliver high baseline accuracy without retrieval, but their marginal benefit from RaR is limited, suggesting they may be more appropriate in settings where resources and latency are less constrained. Clinically fine-tuned models, meanwhile, continue to benefit from RaR, highlighting that retrieval should be viewed as complementary rather than optional.

Thus, the optimal choice of model depends on balancing accuracy needs, interpretability, and resource constraints within the intended clinical context.

Beyond accuracy, our analysis demonstrated that RaR improved factual grounding[6,14] and reduced hallucinations in model outputs. By systematically associating diagnostic responses with specific retrieved content from Radiopaedia.org[19], the framework promoted evidence-based reasoning, which is critical in safety-sensitive applications like radiology. Although clinically relevant evidence was retrieved in less than half of the evaluated cases, most models successfully leveraged this content to produce factually correct responses when it was available. Larger and clinically tuned models demonstrated robustness by correctly responding even when retrieved evidence was irrelevant or insufficient, likely relying on internal knowledge[15]. However, such internally derived answers, while accurate, lack explicit grounding in external sources, raising potential concerns for interpretability and clinical accountability[58]. Smaller models were less resilient when retrieval failed, highlighting their greater reliance on structured external support. Consequently, ensuring high-quality retrieval remains paramount, especially for deployment scenarios where transparency and traceability of decisions are required.

Another noteworthy finding is the relatively frequent occurrence of correct answers despite irrelevant retrieved context. This behavior most likely reflects strong prior knowledge and reasoning capacity in larger and reasoning-optimized models, which can generate accurate responses even when the retrieved evidence is noisy or clinically unhelpful. At the same time, it also indicates retrieval noise or mismatched document selection, where the pipeline surfaces content that is adjacent but not clinically useful. On the one hand, this resilience highlights the capacity of well-trained LLMs to integrate internal knowledge with limited external support[59], a desirable feature when retrieval systems fail. On the other hand, it raises important considerations for interpretability and accountability[60]: correct answers derived without external grounding may be less transparent, harder to audit, and more difficult for clinicians to trust in safety-critical settings. To illustrate this duality, we provide representative examples in **Supplementary Note 2** where models answered correctly despite irrelevant or misleading retrieved excerpts, with annotations showing whether the correctness likely stemmed from internal knowledge or partial overlap with the question. These cases emphasize that retrieval systems play a dual role—not only supplying missing information but also providing traceable evidence that clinicians can verify. Future work should therefore focus on disentangling knowledge-driven versus retrieval-driven correctness, minimizing retrieval noise, and designing systems that can explicitly indicate whether an answer is primarily evidence-grounded or internally derived.

The increased diagnostic reliability introduced by RaR came at a computational cost. Response times significantly increased compared to zero-shot inference due to iterative query refinement, structured evidence gathering, and multi-step coordination. This latency varied substantially by model size and architecture, with smaller models experiencing the largest relative increases, and mid-sized or sparsely activated architectures demonstrating comparatively moderate overhead. Very large models, although capable of achieving high accuracy without retrieval, experienced substantial absolute latency increases without commensurate accuracy gains. Future work should therefore explore optimization strategies to manage computational overhead, such as selective retrieval triggering, parallel evidence pipelines, or methods to distill reasoning into more efficient inference paths.

A related concern is the potential for self-preference bias, since o3 contributed to distractor generation and GPT-4o-mini was used as the orchestration controller in RaR. We emphasize that distractor generation and benchmarking were conducted through fully separated pipelines, and all distractors were systematically reviewed by a board-certified radiologist before inclusion, ensuring that final multiple-choice questions were clinically valid and unbiased. GPT-4o-mini was not evaluated as a question-answering model and played no role in dataset construction or adjudication. Moreover, the multiple-choice framework with human-curated distractors and purely accuracy-based scoring substantially mitigates the risk of self-preference bias, which is more relevant in style-sensitive or evaluator-graded tasks. All models, including those from the GPT family, received identical finalized inputs, and thus operated under the same information constraints. Indeed, recent work suggests that in fact-centric benchmarks with verifiable answers, self-preference effects diminish substantially or align with genuine model superiority[61]. Nevertheless, we acknowledge that future studies could strengthen methodological rigor by ensuring complete model-family independence in dataset construction and orchestration components.

Furthermore, RaR demonstrated value as a decision-support tool for human experts. Providing a board-certified radiologist with the same retrieved context as the RaR system substantially improved their diagnostic accuracy compared to unaided performance. This finding illustrates that the RaR process successfully identified and presented clinically meaningful, decision-relevant evidence that directly supported expert reasoning. The limited number of LLMs significantly outperforming the context-assisted radiologist further underscores the complementary strengths of human expertise and retrieved information. Thus, RaR may serve dual purposes in clinical environments, simultaneously enhancing LLM performance and providing interpretable, actionable evidence to clinicians.

To evaluate whether our findings generalize beyond the RadioRAG benchmark setting, we replicated our analysis on an unseen dataset of radiology board examination questions from a different institution. RaR again improved diagnostic accuracy over zero-shot prompting, preserved factual consistency, and reduced hallucination rates across models, confirming its robustness across settings. However, not all trends reproduced fully. Improvements for mid-sized and clinically fine-tuned models were no longer statistically significant, and the gain from RaR context for the human expert did not reach significance. These discrepancies likely stem from two factors: the smaller sample size of the internal dataset, which reduced statistical power, and the more structured phrasing of board-style questions, which may have facilitated stronger baseline performance for both humans and models. In particular, the higher relevance rate of retrieved evidence in this dataset suggests that the more canonical language of exam-style questions enabled better document matching, narrowing the performance gap between zero-shot and RaR conditions. These findings underscore that while the benefits of RaR broadly generalize, their magnitude may depend on dataset-specific features such as question format and baseline difficulty.

Our study has several important limitations. First, our evaluation relied exclusively on Radiopaedia.org, a trusted, peer-reviewed, and openly accessible radiology knowledge source. We selected Radiopaedia to ensure high-quality and clinically validated content, and we secured explicit approval for its use in this study. While other resources exist, many are either not openly accessible, not peer-reviewed in full, or require separate agreements that were not feasible within the scope of this work. Dependence on a single data provider, however, may restrict retrieval coverage and not capture the full breadth of radiological knowledge. Future studies should aim to incorporate additional authoritative sources, structured knowledge bases, or clinical ontologies to improve coverage and generalizability. Second, although our evaluation spanned two datasets, i.e., (i) the public RadioRAG benchmark (n = 104) and (ii) an independent board-style dataset from the Technical University of Munich (n = 65), the total number of questions remains relatively modest. While both datasets are expert-curated and clinically grounded, larger and more diverse collections encompassing broader clinical scenarios, imaging modalities, and diagnostic challenges are needed to fully assess the robustness and generalizability of RaR. Expanded datasets would also enable higher-powered subgroup analyses and stronger statistical certainty for model- and task-level comparisons. However, creating radiology QA items is highly resource-intensive, requiring significant time and multiple rounds of board-certified radiologist review to ensure that questions are text-based, clinically meaningful, and free from data leakage. To help address this gap, we publicly release our newly developed internal dataset alongside this manuscript, thereby contributing to cumulative dataset growth and enabling future research. Third, the RaR process incurs significant computational overhead, substantially increasing response times compared to conventional zero-shot prompting and traditional single-step RAG. Although response durations remained within feasible limits for non-emergent clinical use cases, the practicality of the proposed method in time-sensitive settings (e.g., acute diagnostic workflows) remains uncertain. Future research should explore optimization techniques, such as parallelization or selective module activation, to mitigate latency without sacrificing diagnostic accuracy or reasoning quality. Fourth, both the RadioRAG and internal board-style datasets consist of static, retrospective QA items that, while clinically representative, do not fully capture the complexity and dynamism of real-world radiology practice. Clinical workflows often involve multimodal inputs (e.g., imaging data, clinical reports), evolving case presentations, and dynamic clinician–AI interactions, none of which are modeled in benchmark-style question formats. Importantly, our study was limited to text-only QA. The multiple-choice format was introduced solely as a benchmarking tool to enable reproducible accuracy measurement across models and humans; in real-world settings, RaR is intended to support open-ended, text-based clinical questions (e.g., "what is the most likely diagnosis given these findings?") rather than exam-style queries. While this design strengthens internal validity, it restricts direct applicability to multimodal radiology tasks. As such, our findings reflect performance in controlled QA environments rather than in prospective or embedded clinical contexts. Future research should therefore validate RaR in real clinical systems, ideally in prospective studies embedded within reporting workflows or decision-support platforms, to assess practical utility, safety, and user impact under real-world conditions. Fifth, despite evaluating a broad range of LLM architectures, parameter scales, and training paradigms, we observed substantial variability in the diagnostic gains attributable to RaR across individual models. This likely reflects a combination of factors, including architectural differences, instruction tuning approaches, and pretraining data composition, as well as implementation-specific elements such as prompt design and module orchestration. Because the RaR pipeline relies on structured prompting and task decomposition, its performance may be sensitive to changes in phrasing, retrieval heuristics, or module coordination. Future work should systematically investigate both model-level and implementation-level sources of variability to develop more robust, generalizable retrieval strategies tailored to different model configurations. Sixth, although the framework improved diagnostic accuracy and factual reliability, it introduced substantial latency overhead. While response durations remained within feasible ranges for non-emergent settings, future research should explore optimization strategies such as asynchronous retrieval, selective triggering of agentic reasoning when model uncertainty is high, and more efficient orchestration of multi-agent pipelines to balance accuracy with computational efficiency.

This study presents a proof-of-concept for a multi-step retrieval and reasoning framework capable of enhancing diagnostic accuracy, factual reliability, and clinical interpretability of LLMs in radiology QA tasks. Our extensive, large-scale analysis of 25 diverse models highlights the complex relationships between retrieval strategy, model scale, and clinical fine-tuning. While RaR shows clear promise, particularly for mid-sized and clinically optimized models, future research is essential to refine retrieval mechanisms, mitigate computational overhead, and validate these systems across broader clinical contexts. As generative AI continues to integrate into medical practice, frameworks emphasizing transparency, evidence-based reasoning, and human-aligned interpretability, such as the RaR approach introduced here, will become increasingly critical for trustworthy and effective clinical decision support. Beyond serving as an automated reasoning pipeline, RaR may also provide a foundation for human–AI collaborative diagnosis. By structuring and externalizing evidence synthesis, the framework enables clinicians to review, validate, and integrate retrieved knowledge into their own diagnostic reasoning. Future iterations of RaR should therefore be explicitly designed to support collaborative workflows, where AI augments rather than replaces clinical expertise, ultimately improving diagnostic confidence, accountability, and patient safety.

# Materials and Methods

## Ethics statement

The methods were performed in accordance with relevant guidelines and regulations. The data utilized in this research was sourced from previously published studies. As the study did not involve human subjects or patients, it was exempt from institutional review board approval and did not require informed consent.

## Dataset

This study utilized two carefully curated datasets specifically designed to evaluate the performance of RaR-powered LLMs in retrieval-augmented radiology QA.

***RadioRAG dataset***

We utilized two previously published datasets from the RadioRAG study[18]: the RSNA-RadioQA[18] and ExtendedQA[18] datasets. The RSNA-RadioQA dataset consists of 80 radiology questions derived from peer-reviewed cases available in the Radiological Society of North America (RSNA) Case Collection. This dataset covers 18 radiologic subspecialties, including breast imaging, chest radiology, gastrointestinal imaging, musculoskeletal imaging, neuroradiology, and pediatric radiology, among others. Each subspecialty contains at least five questions, carefully crafted from clinical histories and imaging descriptions provided in the original RSNA case documentation. Differential diagnoses explicitly listed by original case authors were excluded to avoid biasing model responses. Images were intentionally excluded. Detailed characteristics, including patient demographics and subspecialty distributions, have been previously published and are publicly accessible. The ExtendedQA dataset consists of 24 unique, radiology-specific questions initially developed and validated by board-certified radiologists with substantial diagnostic radiology experience (5–14 years). These questions reflect realistic clinical diagnostic scenarios not previously available online or included in known LLM training datasets. The final RadioRAG dataset used in this study subsequently contains 104 questions combining both RSNA-RadioQA and ExtendedQA.

To ensure consistent evaluation across all models and inference strategies, we applied structured preprocessing to the original RadioRAG dataset, particularly the ExtendedQA portion (n=24), which was initially formatted as open-ended questions. All questions from the RSNA-RadioQA dataset (n=80) were left unchanged. However, for the ExtendedQA subset, each question was first converted into a multiple-choice format while preserving the original stem and correct answer. To standardize the evaluation across both RSNA-RadioQA and ExtendedQA, we then generated three high-quality distractor options for every question in the dataset (n = 104), resulting in a total of four answer choices per item. Distractors were generated using OpenAI's GPT-4o and o3 models, selected for their ability to produce clinically plausible and contextually challenging alternatives. Prompts were designed to elicit difficult distractors, including common misconceptions, closely related entities, or synonyms of the correct answer. All distractors were subsequently reviewed in a structured process by a board-certified radiologist to confirm that they were clinically meaningful, non-trivial, and free of misleading or implausible content. Items failing this review were discarded or revised until they met expert standards. Although o3 and GPT-4o were used to generate preliminary distractors, these were only intermediate drafts. All final multiple-choice options were curated and approved through expert review, ensuring that benchmark items were clinically meaningful, unbiased, and identical across all models irrespective of origin. This hybrid pipeline of LLM-assisted distractor generation plus systematic expert validation has precedent in the educational technology and medical education literature, where it has been shown to produce valid and challenging MCQs when coupled with human oversight[62]. A representative prompt used for distractor generation was:

> *"I have a dataset of radiology questions that are currently open-ended, each with a correct answer provided. I want to transform these into multiple-choice questions (MCQs) by generating four answer options per question (one correct answer + three distractors). The distractors should be plausible and the level of difficulty must be high. If possible, include distractors that are synonyms, closely related concepts, or common misconceptions related to the correct answer."*

**Supplementary Table 1** summarizes the characteristics of the RadioRAG dataset used in this study. The original RSNA-RadioQA questions are publicly available through their original publication[18].

### *Internal generalization dataset*

In addition to the publicly available RadioRAG dataset, we constructed an internal dataset of 65 radiology questions to further evaluate model performance on knowledge domains aligned with German board certification requirements. This dataset was developed and validated by board-certified radiologists (LA with 9 and KB 10 years of clinical experience across subspecialties). Questions were derived from representative diagnostic cases and key concepts covered in the German radiology training curriculum at the Technical University of Munich, ensuring coverage of essential knowledge expected of practicing radiologists in Germany. None of the questions or their formulations are available in online case collections or known LLM training corpora. The internal dataset was formatted as multiple-choice questions following the same pipeline as ExtendedQA. Each question contains 5 options.

# Experimental Design

All retrieval in this study was performed using Radiopaedia.org, a peer-reviewed and openly accessible radiology knowledge base. Radiopaedia was chosen to ensure high-quality and clinically validated content, minimizing the risk of unverified or non-peer-reviewed material. While other authoritative databases exist, many are either not openly available, lack consistent peer review, or require access agreements that were not feasible within the scope of this work. For Radiopaedia, explicit approval for research use was obtained prior to conducting this study.

### *System architecture*

The experimental design centers on an orchestrated retrieval and reasoning framework adapted from LangChain's Open Deep Research pipeline, specifically tailored for radiology QA tasks. As illustrated in **Figure 1**, the pipeline employs a structured, multi-step workflow designed to produce comprehensive, evidence-based diagnostic reports for each multiple-choice question. The reasoning and content-generation process within the RaR orchestration is powered by OpenAI's GPT-4o-mini model, selected for its proficiency in complex reasoning tasks, robust instruction-following, and effective tool utilization. The architecture consists of two specialized modules: (i) a supervisor module and (ii) a research module, coordinated through a stateful directed graph framework. State management within this directed graph framework ensures that all steps in the workflow remain consistent and coordinated. The system maintains a shared memory state, recording the research plan, retrieved evidence, completed drafts, and all module interactions, enabling structured progression from planning through final synthesis. Importantly, GPT-4o-mini functioned only as a fixed orchestration engine coordinating retrieval and structuring evidence; the final diagnostic answer (i.e., the selected option) was always generated by the target model under evaluation. This ensures comparability across models but also clarifies that RaR evaluates how models use structured retrieved evidence rather than their independent ability to perform multi-step reasoning. Because the orchestration process and retrieved context were identical across all tested models, including GPT-family systems, GPT-4o-mini's involvement did not confer any preferential advantage; all models operated under the same inputs and conditions.

### *Preprocessing*

To enable structured, multi-step reasoning in the RaR framework, we implemented a preprocessing step focused on diagnostic abstraction. For each question in the RadioRAG dataset, we used the Mistral Large model to generate a concise, comma-separated summary of key clinical concepts. We selected Mistral Large after preliminary comparisons with alternative LLMs (e.g., GPT-4o-mini, LLaMA-2-70B), as it consistently produced concise, clinically faithful keyword summaries with minimal redundancy, making it particularly well-suited for guiding retrieval (see **Supplementary Note 1** for representative examples). This step was designed to extract the essential diagnostic elements of each question while filtering out rhetorical structure, instructional phrasing (e.g., "What is the most likely diagnosis?"), and other non-clinical language. These keyword summaries served exclusively as internal inputs to guide the RaR system's retrieval process and were not shown to the LLMs as part of the actual question content. The intent was to ensure retrieval was driven by the clinical essence of the question rather than superficial linguistic cues. The prompt used for keyword extraction was:

> *"Extract and summarize the key clinical details from the following radiology question. Provide a concise, comma-separated summary of keywords and key phrases in one sentence only.*
> *Question: {question_text}.*
> *Summary:"*

### *Roles and responsibilities*

The workflow is coordinated primarily by two modules, each with distinct responsibilities: (i) supervisor module and (ii) research module. The supervisor acts as the central orchestrator of the pipeline. Upon receiving a question, the supervisor reviews the diagnostic keywords and multiple-choice options, then formulates a structured research plan dividing the task into clearly defined sections, one for each diagnostic option. This module assigns tasks to individual research modules, each responsible for exploring a single diagnostic choice. Throughout the process, the supervisor ensures strict neutrality, focusing solely on evidence gathering rather than advocating for any particular option. After research modules complete their tasks, the supervisor synthesizes their outputs into a final report, utilizing specialized tools to generate an objective introduction and conclusion.

Each research module independently conducts an in-depth analysis focused on one diagnostic option. Beginning with a clear directive from the supervisor, the research module employs a structured retrieval strategy to obtain relevant evidence. This involves an initial focused query using only essential terms from the diagnostic option, followed by contextual queries combining these terms with clinical features from the question stem (e.g., imaging findings or patient demographics). If retrieval results are inadequate, the module adaptively refines queries by simplifying terms or substituting synonyms. In cases where sufficient evidence is not available after four attempts, the module explicitly documents this limitation. All retrieval tasks utilize Radiopaedia.org exclusively, ensuring clinical accuracy and reliability. After completing retrieval, the research module synthesizes findings into a structured report segment, explicitly highlighting both supporting and contradicting evidence. Each segment includes clearly formatted citations linking directly to source materials, ensuring transparency and verifiability.

### *Retrieval and writing tools*

To facilitate structured retrieval and writing processes, the pipeline utilizes a suite of specialized computational tools dynamically selected based on specific task requirements: (i) search tool, (ii) report structuring tools, and (iii) content generation tool. In the following, details of each tool is explained.

The retrieval mechanism is powered by a custom-built search tool leveraging a locally hosted instance of SearXNG, a privacy-oriented meta-search engine deployed within a containerized Docker environment. This setup ensures consistent and reproducible search results. To maintain quality and clinical reliability, the search tool restricts results exclusively to content from Radiopaedia.org through a two-layer filtering process: first by appending a "site:radiopaedia.org" clause to all queries, and subsequently by performing an explicit domain check on all retrieved results. Raw results are deduplicated and formatted into markdown bundles suitable for seamless integration into subsequent reasoning steps.

The supervisor module employs specific tools to structure the diagnostic report systematically. An initial Sections tool is used to outline the report into distinct diagnostic sections, aligning precisely with the multiple-choice options. Additional specialized tools generate standardized Introduction and Conclusion sections: the Introduction tool summarizes essential clinical details from the question, and the Conclusion tool objectively synthesizes findings from all diagnostic sections, emphasizing comparative diagnostic considerations without bias.

The research module utilizes a dedicated Section writing tool to construct standardized report segments. Each segment begins with a concise synthesis of retrieved evidence, followed by interpretive summaries clearly identifying points supporting and contradicting each diagnostic choice. Citations are integrated inline, referencing specific Radiopaedia[19] URLs for traceability.

### *Report assembly and persistence*

Upon completion of individual research segments, the supervisor module compiles the final diagnostic report, verifying the completeness and quality of all sections. The resulting structured report, including introduction, detailed analysis of diagnostic options, and conclusion, is then immediately persisted in a robust manner. Reports are streamed incrementally into newline-delimited JSON (NDJSON) format, preventing data loss in case of interruptions. This storage method supports efficient resumption by checking previously completed entries, thus avoiding redundant processing. After processing all questions within a given batch, individual NDJSON entries are consolidated into a single comprehensive JSON file, facilitating downstream analysis and evaluation.

## Baseline comparison systems

Each model was evaluated under three configurations: (i) zero-shot prompting (conventional QA), (ii) conventional online RAG[18], and (iii) our proposed RaR framework.

### *Baseline 1: Zero-shot prompting pipeline*

In the zero-shot prompting baseline, models received no external retrieval assistance or context. Instead, each model was presented solely with the multiple-choice questions from the RadioRAG dataset (question stem and four diagnostic options) and prompted to select the correct answer based entirely on their pre-trained knowledge. Models generated their responses autonomously without iterative feedback, reasoning prompts, or additional information.

The exact standardized prompt used for this configuration is provided below:

> *"You are a highly knowledgeable medical expert. Below is a multiple-choice radiology question. Read the question carefully. Provide the correct answer by selecting the most appropriate option from A, B, C, or D.*
> *Question:*
> *{question}*
>
> *Options:*
> *{options}"*

### *Baseline 2: Conventional online RAG pipeline*

The conventional online RAG baseline was implemented following a state-of-the-art non-agentic retrieval framework previously developed for radiology question answering by Tayebi Arasteh et al[18]. The system employs GPT-3.5-turbo to automatically extract up to five representative radiology keywords from each question, optimized experimentally to balance retrieval quality and efficiency. These keywords were used to retrieve relevant articles from Radiopaedia.org, with each article segmented into overlapping chunks of 1,000 tokens. Chunks were then converted into vector embeddings (OpenAI's text-embedding-ada-002) and stored in a temporary vector database. Subsequently, the embedded original question was compared against this database to retrieve the top three matching text chunks based on cosine similarity. These retrieved chunks served as external context provided to each LLM alongside the original multiple-choice question. Models were then instructed to answer concisely based solely on this context, explicitly stating if the answer was unknown.

The exact standardized prompt used for this configuration is provided below:

> *"You are a highly knowledgeable medical expert. Below is a multiple-choice radiology question accompanied by relevant context (report). First, read the report, and then the question carefully. Use the retrieved context to answer the question by selecting the most appropriate option from A, B, C, or D. Otherwise, if you don't know the answer, just say that you don't know.*
> *Report:*
> *{report}*
>
> *Question:*
> *{question}*
>
> *Options:*
> *{options}"*

# Evaluation

SW, JS, TTN, and STA performed model evaluations. We assessed both small and large-scale LLMs using responses generated between July 1 – August 22, 2025. For each of the 104 questions in the RadioRAG benchmark dataset, as well as each of the 65 questions in the unseen generalization dataset, models were integrated into a unified evaluation pipeline to ensure consistent testing conditions across all settings. The evaluation included 25 LLMs: Ministral-8B, Mistral Large, Llama3.3-8B[45,46], Llama3.3-70B[45,46], Llama3-Med42-8B[43], Llama3-Med42-70B[43], Llama4 Scout 16E[33], DeepSeek R1-70B[44], DeepSeek-R1[44], DeepSeek-V3[47], Qwen 2.5-0.5B[41], Qwen 2.5-3B[41], Qwen 2.5-7B[41], Qwen 2.5-14B[41], Qwen 2.5-70B[41], Qwen 3-8B[48], Qwen 3-235B[48], GPT-3.5-turbo, GPT-4-turbo[8], o3, GPT-5[49], MedGemma-4B-it[42], MedGemma-27B-text-it[42], Gemma-3-4B-it[50,51], and Gemma-3-27B-it[50,51]. These models span a broad range of parameter scales (from 0.5B to over 670B), training paradigms (instruction-tuned, reasoning-optimized, clinically aligned, and general-purpose), and access models (open-source, open-weights, or proprietary). They also reflect architectural diversity, including dense transformers and MoE[52] systems. Full model specifications, including size, category, accessibility, knowledge cutoff date, context length, and developer are provided in **Table 1**. For clarity, GPT-5 is included here as a widely used system-level benchmark. As noted in OpenAI's documentation, GPT-5 internally routes queries across different underlying models depending on the task, and should therefore be regarded as a system rather than a fixed architecture. All models were run with deterministic decoding parameters (temperature = 0, top-p = 1, no top-k or nucleus sampling). No random seeds or stochastic ensembles were used, and each model produced a single, reproducible response per question. This ensured that performance differences reflected model reasoning ability rather than variability introduced by random sampling.

## *Accuracy assessment*

Accuracy was determined by comparing each LLM's response to the correct option. For this, we used Mistral Large as an automated adjudicator. Importantly, this role was not generative reasoning but a constrained verification task: the adjudicator only needed to check whether the correct option appeared in the response. This made the process essentially binary option-matching rather than open-ended judgment, thereby minimizing any risk of hallucination.

Constrained decoding was not applied during answer generation; instead, each model was explicitly prompted to select one option (A–D for the RadioRAG benchmark, or A–E for the internal dataset). In rare cases where a model output included multiple options (e.g., "A and D"), scoring was based strictly on whether the correct option appeared explicitly and unambiguously in the response. If the correct option was included, the response was counted as correct; otherwise, it was scored as incorrect. This ensured reproducibility and avoided bias across models.

For each multiple-choice question, both the LLM's response and the correct answer (including its corresponding letter and option) were provided to Mistral Large via a standardized prompt. The adjudicator was explicitly instructed to respond only with "Yes" if the correct answer was present, or "No" otherwise, ensuring that outputs were strictly bounded and reproducible. A "Yes" was scored as 1 (correct), and a "No" was scored as 0 (incorrect), ensuring a consistent and unbiased measure of diagnostic accuracy.

The exact standardized prompt used for this configuration is provided below:

> *"You are a highly knowledgeable medical expert. Determine whether the Correct Answer appears within the LLMs response, fully or as a clear part of the explanation, even if the wording differs. Respond with 'Yes' if the Correct Answer can be found in the LLMs response; otherwise respond with 'No'.*
>
> *LLMs response:*
> *{llms_response}*
>
> *Correct Answer:*
> *{correct_answer}"*

To validate this automated procedure, we manually reviewed question outputs for a representative subset of LLMs and confirmed complete concordance with the adjudicator's results.

*Factuality assessment*

To evaluate the factual reliability of model outputs under the RaR framework, we conducted a targeted hallucination analysis across all 104 questions in the RadioRAG benchmark[18] (and separately across all 65 questions in the unseen generalization dataset). This analysis aimed to differentiate model errors due to flawed reasoning from those caused by insufficient or irrelevant evidence, and to assess the extent to which final answers were grounded in the retrieved context.

Each RaR response was reviewed by a board-certified radiologist (TTN) with seven years of experience in diagnostic and interventional radiology. For every question, the following three criteria were assessed: (i) whether the retrieved Radiopaedia context was clinically relevant to the question, (ii) whether the model's final answer was consistent with that context, and (iii) whether the final answer was factually correct.

Context was classified as clinically relevant only if it contained no incorrect or off-topic content with respect to the diagnostic question. This strict definition ensured that relevance was not based on superficial keyword overlap but on the actual clinical utility of the content. Retrievals were deemed relevant only when the retrieved material included appropriate imaging findings, clinical clues, or differential diagnoses applicable to the question stem.

Hallucinations were defined as cases in which the model produced an incorrect answer despite being provided with clinically relevant context. These represent failures of reasoning or synthesis rather than of retrieval. Given the high-stakes nature of radiologic diagnosis, identifying such errors is essential for understanding model reliability and safety.

We also documented instances where models answered questions correctly despite being supplied with irrelevant or unhelpful context. These "correct despite irrelevant context" cases reflect scenarios in which the model relied on internal knowledge rather than external grounding. While not classified as hallucinations, these responses raise questions about the transparency, traceability, and consistency of model behavior in the absence of meaningful retrieval.

*Time analysis*

To evaluate the computational cost associated with RaR, we measured per-question response times for both zero-shot prompting and the RaR framework using the 104-question RadioRAG benchmark (and separately using the 65 questions of the unseen generalization dataset). Timing logs were collected from structured output directories for each model. For each dataset, we measured a fixed initialization overhead corresponding to the context construction phase unique to RaR inference. On the RadioRAG dataset (n = 104), this overhead averaged 10,554.6 seconds per model (≈101.5 seconds per question). On the internal dataset (n = 65), the overhead averaged 5,754.9 seconds per model (≈88.5 seconds per question). Together, this corresponds to a total of 16,301 seconds across both datasets, or ≈97 seconds per question on average. These overheads were distributed uniformly across all questions to ensure fair per-question latency estimates.

To ensure robust comparison and mitigate the influence of extreme values, outlier durations were handled using the Tukey method[63]. Specifically, any response time that exceeded the typical upper range, defined as values greater than the third quartile by more than 1.5 times the interquartile range, was considered an outlier and replaced with the mean of the remaining non-outlier values for that model and inference strategy. For each model, we computed the mean and standard deviation of response times under both conditions. Additionally, we calculated the absolute difference in average response time per question and the relative increase, defined as the ratio of mean RaR response time to mean zero-shot response time.

To contextualize timing behavior across a heterogeneous model set, we grouped models according to both parameter scale and architectural characteristics. This grouping approach reflected the practical computational load of each model more accurately than parameter count alone. Six distinct groups were defined: (i) the DeepSeek MoE group, including DeepSeek-R1 and DeepSeek-V3; (ii) the large model group (120–250 billion parameters), including Qwen 3-235B, Mistral Large, and Llama4 Scout 16E; (iii) the medium-scale group (~70B), comprising DeepSeek R1-70B, Llama3.3-70B, Qwen2.5-70B, and Llama3-Med42-70B; (iv) the Gemma-27B group, containing Gemma-3-27B-it and MedGemma-27B-text-it; (v) the small model group (7–8B), including Qwen 2.5-70B, Qwen3-8B, Llama3-Med42-8B, Llama3.3-8B, and Ministral-8B; and (vi) the mini model group (3–4B), consisting of Gemma-3-4B-it, MedGemma-4B-it, and Qwen 2.5-3B. Group-level averages and standard deviations were calculated across constituent models and are reported in **Table 4**.

All timing evaluations were performed under identical system conditions to ensure fair comparisons. While absolute response times may vary with hardware and load, the relative increases provide a stable and interpretable metric for assessing the computational implications of RaR.

## *Human evaluation*

To benchmark LLM performance against domain expertise, we conducted a human evaluation involving a board-certified radiologist (TTN) with seven years of experience in diagnostic and interventional radiology. The evaluation followed a two-phase design to mirror the LLM configurations.

In the first phase, the radiologist answered all 104 questions from the RadioRAG benchmark (and separately all 65 questions from the internal generalization dataset) without any external assistance, analogous to zero-shot prompting. The expert was fully blinded to the LLM responses, dataset construction process, and reference standard answers, which remained inaccessible throughout the entire study, including after task completion and up to manuscript submission. Responses were recorded as final, and no additional time or information resources were permitted during this phase.

In the second phase, we aimed to isolate the contribution of the RaR component, independent of generative reasoning. For this, the same radiologist was provided with the contextual evidence retrieved by the RaR system for each question, the same Radiopaedia excerpts that were used as inputs for RaR-powered LLM inference. The radiologist completed this phase only after finishing the unaided zero-shot phase, and did not have access to the correct answers or to their own previous responses, thereby avoiding bias from prior knowledge. The radiologist answered the same 104 questions again (and separately the same 65 questions of the internal generalization dataset), this time using the retrieved context as decision support, without access to the original question-answer pairs or their previous responses. The format and presentation of the contextual evidence were identical to what the LLMs received during RaR-powered inference, ensuring comparability.

This design enabled us to disentangle the effects of information retrieval from language model reasoning, by comparing unaided radiologist performance, radiologist performance with context, and RaR-based LLM outputs under standardized conditions. Accuracy was computed using the same evaluation criteria applied to LLMs. Statistical comparisons between human and model responses were performed using McNemar's test on paired question-level outcomes. Confidence intervals and p-values were adjusted for multiple comparisons using the false discovery rate.

## Statistical analysis

Statistical analysis was performed using Python v3.11 with SciPy v1.10, NumPy v1.25.2, and statsmodels v0.14.5 packages. For each dataset, bootstrapping with 1,000 redraws was used to estimate means, standard deviations, and 95% confidence intervals (CI)[64]. A strictly paired design ensured identical redraws across conditions[65]. To assess statistical significance of individual model-level comparisons between inference strategies, exact McNemar's test[66] (based on the binomial distribution) was applied to each model separately on paired question-level outcomes. Resulting p-values were corrected for multiple comparisons using the false discovery rate, with a significance threshold of 0.05. These values are reported in **Table 2** and per-model Results subsections. For group-level comparisons (e.g., zero-shot vs. RaR across mid-sized models), paired two-tailed t-tests were used to compare average accuracy across all models in the group. These p-values therefore reflect differences at the cohort level rather than for any single model, and are explicitly labeled as such in the Results. To explore the relationship between model size and performance, Pearson correlation coefficients were computed between parameter counts and accuracy values within the Qwen 2.5 model family, separately for each inference strategy.

## Data availability

All data in this study are available. The RadioRAG dataset including the original RSNA-RadioQA and ExtendedQA are available via the original RadioRAG publication[18].

## Code availability and reproducibility

All source code, configurations, and parameters used in this work are publicly available. The RaR pipeline, developed in Python 3.11, is available at: https://github.com/sopajeta/RaR. Our implementation relies on several key frameworks and tools. We used LangChain Open Deep Research (https://github.com/langchain-ai/deep-research) for experimental modules, LangChain v0.3.25 (https://github.com/langchain-ai/langchain) for orchestration and management, and LangGraph v0.4.1 (https://github.com/langchain-ai/langgraph) to support multi-step control flow and task decomposition. Model access and embedding generation were handled via the OpenAI Python SDK v1.77.0 (https://platform.openai.com). The SearxNG metasearch engine (https://github.com/searxng/searxng) was also deployed via Docker v25.0.2 (https://www.docker.com) and used for online web retrieval.

The traditional online RAG pipeline is hosted at https://github.com/tayebiarasteh/RadioRAG, which relies on the LangChain v0.1.0, Chroma (https://www.trychroma.com) for vector storage, and the OpenAI API v1.12 for embeddings.

All locally deployed language models sourced from Hugging Face, were assessed and used between July 1 – August 22, 2025, and are explicitly listed below, with corresponding URLs:

- Qwen 2.5-0.5B: https://huggingface.co/Qwen/Qwen2.5-0.5B
- Qwen 2.5-3B: https://huggingface.co/Qwen/Qwen2.5-3B
- Qwen 2.5-7B: https://huggingface.co/Qwen/Qwen2.5-7B
- Qwen 2.5-14B: https://huggingface.co/Qwen/Qwen2.5-14B
- Qwen 2.5-70B: https://huggingface.co/Qwen/Qwen2.5-72B
- Qwen 3-8B: https://huggingface.co/Qwen/Qwen3-8B
- Qwen 3-235B: https://huggingface.co/Qwen/Qwen3-235B-A22B
- Llama 3.3-8B: https://huggingface.co/meta-llama/Meta-Llama-3-8B
- Llama 3.3-70B: https://huggingface.co/meta-llama/Llama-3.3-70B-Instruct
- Llama 3-Med42-70B: https://huggingface.co/m42-health/Llama3-Med42-70B
- Llama 3-Med42-8B: https://huggingface.co/m42-health/Llama3-Med42-8B
- Llama4 Scout 16E: https://huggingface.co/meta-llama/Llama-4-Scout-17B-16E
- Mistral Large: https://huggingface.co/mistralai/Mistral-Large-Instruct-2407
- Ministral 8B: https://huggingface.co/mistralai/Ministral-8B-Instruct-2410
- Gemma-3-4B-it: https://huggingface.co/google/gemma-3-4b-it
- Gemma-3-27B-it: https://huggingface.co/google/gemma-3-27b-it
- Medgemma-4B-it: https://huggingface.co/google/medgemma-4b-it
- Medgemma-27B-text-it: https://huggingface.co/google/medgemma-27b-text-it
- DeepSeek-V3: https://huggingface.co/deepseek-ai/DeepSeek-V3
- DeepSeek-R1: https://huggingface.co/deepseek-ai/DeepSeek-R1
- DeepSeek-R1-70B: https://huggingface.co/deepseek-ai/DeepSeek-R1-Distill-Llama-70B

All the previously mentioned LLMs were served using vLLM v0.9.0 (https://github.com/vllm-project/vllm) with tensor parallelism set to the number of GPUs inside the node, except for models under 3 billion parameters, which were served without tensor parallelism.

All OpenAI-hosted models were accessed through direct REST API calls to the OpenAI endpoints (https://platform.openai.com). The exact versions employed in this study are as follows:
- GPT-5 (2025-08-07)

- O3 (2025-04-16)
- GPT-4-Turbo (2024-04-09)
- GPT-3.5-Turbo (2024-01-25)

# Hardware

For the majority of experiments, particularly those involving standard LLMs, the computations were performed on GPU nodes equipped with Nvidia H100 and H200 accelerators. The H100 configuration consisted of four Nvidia H100 GPUs, each providing 94 GB of HBM2e memory and operating at a 500 W power limit. These GPUs were paired with two AMD EPYC 9554 "Genoa" processors based on the Zen 4 architecture, each offering 64 high-performance cores running at 3.1 GHz. The H200 configuration featured four Nvidia H200 GPUs, each offering 141 GB of high-bandwidth memory also at 500 W, coupled to the same dual AMD EPYC 9554 processor configuration. This combination of high-end Nvidia accelerators from NHR@FAU's Helma Cluster (https://doc.nhr.fau.de/clusters/helma/) provided the necessary computational capabilities for inferencing the majority of the LLMs used during our experiments.

Experiments involving extremely large-scale architectures, such as the DeepSeek R1 or V3 model and other similarly demanding workloads, were executed on nodes equipped with AMD's MI300-series accelerators. In these cases, the MI300X configuration was utilized, which combined a dual-socket AMD EPYC 9474F platform with a total of 96 CPU cores and 2304 GB of DDR5-5600 system memory, together with eight AMD Instinct MI300X accelerators. Each MI300X GPU offered 192 GB of memory, enabling inference runs that required massive parameter counts and exceptional memory capacity (Deepseek R1 with 671 billion parameters). Additional experimentation also leveraged AMD Instinct MI300A nodes that integrate 24-core CPUs with unified on-package memory, with a total of 512 GB shared across four accelerators. The hardware used in our experiments included a local machine with an Intel Pentium CPU with 2 cores and 8 GB Memory for consuming API endpoints.

# Funding

This research is supported by BayernKI, the central infrastructure for the State of Bavaria to advance academic AI research. The authors gratefully acknowledge the HPC resources provided by the Erlangen National High Performance Computing Center (NHR@FAU) of the Friedrich-Alexander-Universität Erlangen-Nürnberg. NHR funding is provided by federal and Bavarian state authorities. NHR@FAU hardware is partially funded by the Deutsche Forschungsgemeinschaft (DFG) – 440719683. DT was supported by grants from the DFG (NE 2136/3-1, LI3893/6-1, TR 1700/7-1) and is supported by the German Federal Ministry of Education (TRANSFORM LIVER, 031L0312A; SWAG, 01KD2215B) and the European Union's Horizon Europe and innovation programme (ODELIA [Open Consortium for Decentralized Medical Artificial Intelligence], 101057091). KB received grants from the European Union (101079894), Bayern Innovativ,

## Author contributions

The formal analysis was conducted by SW, JS, and STA. The original draft was written by STA, JS, and SW and edited by STA. JS developed the codes for analysis and pipeline; SW configured and maintained the LLM-serving infrastructure. The experiments were performed by SW and JS. The statistical analyses were performed by SW, JS, and STA. The internal dataset was curated by LA and KB. DT, TTN, KB, LA, MR, and STA provided clinical expertise. SW, JS, DT, ML, TTN, KB, MR, HK, GW, AM, and STA provided technical expertise. The study was defined by STA. All authors read the manuscript and agreed to the submission of this paper.

## Competing interests

SW is partially employed by DATEV eG, Germany. DT received honoraria for lectures by Bayer, GE, Roche, AstraZeneca, and Philips and holds shares in StratifAI GmbH, Germany, and in Synagen GmbH, Germany. ML is employed by Generali Deutschland Services GmbH, Germany and is an editorial board at European Radiology Experimental. KB and LA are trainee editorial boards at Radiology: Artificial Intelligence. AM is an associate editor at IEEE Transactions on Medical Imaging. STA is an editorial board at Communications Medicine and at European Radiology Experimental, a trainee editorial board at Radiology: Artificial Intelligence, and is partially employed by Synagen GmbH, Germany. The other authors do not have any competing interests to disclose.

# Supplementary information

**Supplementary Table 1: Characteristics of the RadioRAG dataset used in this study.** The RadioRAG dataset combines RSNA-RadioQA and ExtendedQA, as introduced in the original RadioRAG study. Patient demographic information (age and sex) is based solely on the RSNA-RadioQA subset due to missing metadata in ExtendedQA. Each question may be assigned to multiple radiology subspecialties. *Age and sex statistics reflect only the RSNA-RadioQA subset. *Youngest patient was 2 days old. SD: Standard deviation; N/A: Not available.

| Value | RadioRAG dataset |
|---|---|
| Patient age [years]* <br> Median <br> Mean ± SD <br> Range | 44 <br> 44 ± 21 <br> (0**, 80) |
| Patient sex [n (%)]* <br> Total <br> Female <br> Male | 80 (100%) <br> 37 (46%) <br> 43 (54%) |
| Number of questions per subspecialty [n (%)] | |
| Total | 104 (100%) |
| Breast Imaging | 10 (10%) |
| Cardiac | 10 (10%) |
| Chest | 20 (19%) |
| CT | 35 (34%) |
| Emergency Radiology | 9 (9%) |
| Gastrointestinal | 18 (17%) |
| Genitourinary | 9 (9%) |
| Head and Neck | 10 (10%) |
| MRI | 27 (26%) |
| Molecular Imaging | 11 (11%) |
| Musculoskeletal | 20 (19%) |
| Neuroradiology | 11 (11%) |
| Nuclear Medicine | 13 (12%) |
| Oncologic Imaging | 16 (15%) |
| Pediatric | 8 (8%) |
| Radiation Oncology | 9 (9%) |
| Ultrasound | 10 (10%) |
| Vascular Imaging | 16 (15%) |

**Supplementary Table 2: Accuracy of 25 language models on the RadioRAG dataset restricted to items with clinically relevant retrieved context.** Results are shown for the 48 of 104 questions (46 %) in which the retrieved evidence was judged relevant by expert review (same relevance labels as in **Table 3**). For each model, accuracies are reported for zero-shot prompting and for the RaR framework. Values represent bootstrap means ± standard deviations with 95 % percentile confidence intervals, based on 1,000 shared resampling draws across models and strategies. "Total correct" indicates the number of correct answers out of 48. P-values compare RaR and zero-shot results on paired items using the exact two-sided McNemar test; multiple comparisons were adjusted using FDR.

| Model name | Zero-shot | | | RaR | |
|---|---|---|---|---|---|
| | Accuracy (%) | Total correct (n) | P-value | Accuracy (%) | Total correct (n) |
| Ministral-8B | 48 ± 7 [35, 62] | 23 | 0.122 | 69 ± 7 [56, 81] | 33 |
| Mistral Large (123B) | 71 ± 7 [56, 83] | 34 | 0.122 | 88 ± 5 [77, 96] | 42 |
| Llama3.3-8B | 66 ± 7 [52, 79] | 32 | 0.806 | 63% ± 7 [48, 77] | 30 |
| Llama3.3-70B | 81 ± 6 [71, 92] | 39 | 0.566 | 88 ± 5 [77, 96] | 42 |
| Llama3-Med42-8B | 68 ± 7 [54, 81] | 33 | 0.539 | 77 ± 6 [65, 90] | 37 |
| Llama3-Med42-70B | 71 ± 7 [56, 83] | 34 | 0.122 | 86 ± 5 [75, 94] | 41 |
| Llama4 Scout 16E | 77 ± 6 [65, 88] | 37 | 0.146 | 90 ± 5 [79, 98] | 43 |
| DeepSeek R1-70B | 86 ± 5 [75, 94] | 41 | 0.748 | 90 ± 5 [79, 98] | 43 |
| DeepSeek R1 (671B) | 90 ± 5 [79, 98] | 43 | 0.748 | 94 ± 3 [85, 100] | 45 |
| DeepSeek-V3 (671B) | 77 ± 6 [65, 88] | 37 | 0.122 | 92 ± 4 [83, 98] | 44 |
| Qwen 2.5-0.5B | 41 ± 7 [29, 54] | 20 | 0.999 | 44 ± 7 [31, 58] | 21 |
| Qwen 2.5-3B | 56 ± 7 [44, 71] | 27 | 0.146 | 71 ± 6 [58, 83] | 34 |
| Qwen 2.5-7B | 54 ± 7 [40, 69] | 26 | 0.122 | 75 ± 6 [62, 85] | 36 |
| Qwen 2.5-14B | 60 ± 7 [46, 75] | 29 | 0.122 | 79 ± 6 [67, 90] | 38 |
| Qwen 2.5-70B | 75 ± 6 [62, 88] | 36 | 0.146 | 90 ± 4 [81, 98] | 43 |
| Qwen 3-8B | 71 ± 6 [58, 83] | 34 | 0.146 | 88 ± 5 [77, 96] | 42 |
| Qwen 3-235B | 83 ± 5 [73, 94] | 40 | 0.566 | 90 ± 5 [81, 98] | 43 |
| GPT-3.5-turbo | 54 ± 7 [40, 69] | 26 | 0.148 | 71 ± 7 [56, 83] | 34 |
| GPT-4-turbo | 73 ± 7 [58, 85] | 35 | 0.506 | 81 ± 6 [69, 92] | 39 |
| o3 | 90 ± 4 [81, 98] | 43 | 0.391 | 96 ± 3 [90, 100] | 46 |
| GPT-5 | 90 ± 4 [79, 98] | 43 | 0.595 | 94 ± 4 [85, 100] | 45 |
| MedGemma-4B-it | 50 ± 7 [35, 65] | 24 | 0.391 | 62 ± 7 [48, 77] | 30 |
| MedGemma-27B-text-it | 73 ± 6 [60, 85] | 35 | 0.122 | 94 ± 3 [85, 100] | 45 |
| Gemma-3-4B-it | 44 ± 7 [29, 58] | 21 | 0.391 | 56 ± 7 [42, 69] | 27 |
| Gemma-3-27B-it | 64 ± 7 [50, 77] | 31 | 0.122 | 86 ± 5 [75, 94] | 41 |

**Supplementary Table 3**: **RaR-powered responses and retrieved sources for GPT-3.5-turbo on two RSNA-RadioQA questions.** Full answers generated by GPT-3.5-turbo using the RaR framework for the first two questions of the RSNA-RadioQA dataset, along with the most relevant sources retrieved from Radiopaedia. Responses illustrate the model's reasoning and evidence grounding within the RaR pipeline.

| Question identifier | LLM's (GPT-3.5-turbo) answer | Most relevant sources |
|---|---|---|
| RSNA-RadioQA-Q1 | A: BI-RADS 4 - Suspicious abnormality requiring biopsy: BI-RADS 4 lesions are characterized by suspicious findings with an intermediate risk of malignancy. The presence of rim calcifications within a fat-containing mass raises concern for malignancy. Biopsy is generally recommended for BI-RADS 4 lesions, especially if calcifications are suspicious. B: BI-RADS 2 - Rim Calcifications: Rim calcifications in fat-containing masses are typically benign and classified as BI-RADS 2. These calcifications are often associated with benign processes like fat necrosis or oil cysts. The imaging features of a round, circumscribed fat-containing mass with rim calcifications align with a benign diagnosis. C: BI-RADS 3 - Probably benign finding: BI-RADS 3 category is designated for probably benign lesions with a low likelihood of malignancy. Fat-containing lesions with rim calcifications may fall into this category if they demonstrate typical benign features. Short-interval follow-up is recommended for BI-RADS 3 lesions to monitor for any interval changes. D: Oil cyst secondary to fat necrosis: Oil cysts are benign lesions resulting from fat necrosis, often following trauma or surgery. These cysts typically present as fat-containing masses with rim calcifications, consistent with the imaging findings described. The absence of a clear history of trauma or surgery does not exclude the possibility of an oil cyst. Conclude with a Final Answer: Based on the detailed analysis of the imaging findings and the differential diagnosis provided in the report, the most likely diagnosis for the 66-year-old female with a palpable right breast lump at 10 o'clock, characterized by a round, circumscribed fat-containing mass with rim calcifications on mammography, is: D: Oil cyst secondary to fat necrosis. Reasoning: The presence of a round, circumscribed fat-containing mass with rim calcifications in the absence of a clear history of trauma or surgery aligns with the typical imaging features of an oil cyst secondary to fat necrosis. Oil cysts are benign lesions associated with fat necrosis and are characterized by the presence of fat within the lesion and rim calcifications on mammography. This diagnosis is further supported by the benign nature of rim calcifications in fat-containing masses, typically classified as BI-RADS 2. Therefore, considering the imaging findings and clinical context, the most likely explanation for the palpable abnormality in this case is an oil cyst secondary to fat necrosis. | https://radiopaedia.org/articles/breast-imaging-reporting-and-data-system-bi-rads-assessment-category-4, https://radiopaedia.org/cases/suspicious-breast-calcifications-bi-rads-4, https://radiopaedia.org/articles/fat-containing-breast-lesions, https://radiopaedia.org/articles/breast-calcifications, https://radiopaedia.org/cases/screening-mammography-rim-calcifications, https://radiopaedia.org/articles/breast-imaging-reporting-and-data-system-bi-rads-assessment-category-2, https://radiopaedia.org/cases/breast-calcification-bi-rads-2, https://radiopaedia.org/articles/breast-imaging-reporting-and-data-system-bi-rads-assessment-category-3, https://radiopaedia.org/articles/fat-necrosis-breast-2, https://radiopaedia.org/articles/breast-imaging-reporting-and-data-system-bi-rads-2, https://radiopaedia.org/articles/oil-cyst-breast |
| RSNA-RadioQA-Q2 | A: Low-grade fibrosarcoma: Low-grade fibrosarcoma is a rare malignant mesenchymal tumor characterized by spindle cell morphology with a herringbone pattern. Immunohistochemical markers for fibrosarcoma often include vimentin positivity, but specific markers can vary. Beta-catenin nuclear staining is typically absent in fibrosarcomas, which helps differentiate them from other entities. The absence of nuclear beta-catenin staining in the presented case suggests low-grade fibrosarcoma is less likely. B: Phyllodes tumor: Phyllodes tumors are rare fibroepithelial breast tumors with a biphasic pattern of epithelial and spindle cell stromal components. Beta-catenin staining may show nuclear positivity in some cases, reflecting Wnt pathway activation. The clinical presentation of a rapidly growing breast mass in adult women aligns with the patient's characteristics. The presence of spindle cells with nuclear beta-catenin staining in the presented case supports the possibility of a phyllodes tumor. C: Metaplastic carcinoma: Metaplastic carcinoma is a rare primary breast malignancy with both epithelial and mesenchymal components.Beta-catenin expression patterns may show aberrant nuclear localization, associated with Wnt signaling pathway activation. The patient's history of prior radiation exposure aligns with a potential risk factor for metaplastic carcinoma.\n- The spindle cell morphology and nuclear beta-catenin staining in the presented case could be consistent with metaplastic carcinoma. D: Breast fibromatosis: Breast fibromatosis is a rare benign but locally aggressive tumor characterized by spindle cell proliferation and strong nuclear beta-catenin staining.Clinical association with prior radiation exposure matches the patient's history. The spindle cell morphology and nuclear beta-catenin staining in the presented case are in line with breast fibromatosis. Final Answer: D: Breast fibromatosis. Summary of Reasoning: Considering the patient's high risk of breast malignancy due to BRIP1 mutation and prior radiation exposure, the presence of spindle cells with nuclear beta-catenin staining in the breast mass suggests a diagnosis of breast fibromatosis. This rare tumor entity aligns with the clinical presentation, histopathological findings, and the patient's medical history, making it the most likely diagnosis in this case. | https://radiopaedia.org/articles/fibrosarcoma-of-the-breast, https://radiopaedia.org/cases/breast-sarcoma-2?lang=gb, https://radiopaedia.org/articles/phyllodes-tumour-1, https://radiopaedia.org/articles/metaplastic-breast-carcinoma, https://radiopaedia.org/cases/metaplastic-carcinoma-breast, https://radiopaedia.org/articles/fibromatosis-of-the-breast, https://radiopaedia.org/cases/desmoid-fibromatosis-of-the-breast |

**Supplementary Table 4**: **Zero-shot responses from GPT-3.5-turbo on the first 20 questions of the RSNA-RadioQA dataset.** Model-generated answers are shown without RaR. Responses reflect zero-shot inference using only the question text as input.

| Question identifier | LLM's (GPT-3.5-turbo) answer |
|---|---|
| RSNA-RadioQA-Q1 | The most likely diagnosis is a benign breast lesion, such as a lipoma or fibroadenoma, with associated calcifications. |
| RSNA-RadioQA-Q2 | The most likely diagnosis is desmoid tumor (also known as aggressive fibromatosis). |
| RSNA-RadioQA-Q3 | The most likely diagnosis is Cat Scratch Disease (CSD) with suppurative granulomatous lymphadenitis. |
| RSNA-RadioQA-Q4 | The most likely diagnosis is bilateral invasive ductal carcinoma of the breast, ER positive, PR positive, and HER2 negative, with no lymph node involvement. |
| RSNA-RadioQA-Q5 | The most likely diagnosis is pulmonary rhabdomyosarcoma. |
| RSNA-RadioQA-Q6 | The most likely diagnosis is severe COVID-19 pneumonia. |
| RSNA-RadioQA-Q7 | The most likely diagnosis is pulmonary lymphangioleiomyomatosis (LAM), a characteristic manifestation of Tuberous Sclerosis Complex (TSC). |
| RSNA-RadioQA-Q8 | The most likely diagnosis is a patellar fracture. |
| RSNA-RadioQA-Q9 | The most likely diagnosis is diverticulitis with associated pericolic inflammation and phlegmon formation. |
| RSNA-RadioQA-Q10 | The most likely diagnosis is Hashimoto's thyroiditis. |
| RSNA-RadioQA-Q11 | The most likely diagnosis is orbital sarcoidosis. |
| RSNA-RadioQA-Q12 | The most likely diagnosis is epidural spinal hemangioma. |
| RSNA-RadioQA-Q13 | The most likely diagnosis is a stress fracture of the left femoral neck. |
| RSNA-RadioQA-Q14 | The most likely diagnosis is Kienböck's disease. |
| RSNA-RadioQA-Q15 | The most likely diagnosis is a benign simple cyst of the liver. |
| RSNA-RadioQA-Q16 | The most likely diagnosis is Giant Cell Tumor of Tendon Sheath (GCTTS). |
| RSNA-RadioQA-Q17 | The most likely diagnosis is Alveolar soft part sarcoma (ASPS). |
| RSNA-RadioQA-Q18 | The most likely diagnosis is patellar tendon avulsion fracture. |
| RSNA-RadioQA-Q19 | The most likely diagnosis is benign complicated cysts, given the resolution of the mass with aspiration and the benign nature of the identified cysts on imaging. |
| RSNA-RadioQA-Q20 | The most likely diagnosis is a retroperitoneal teratoma. |

**Supplementary Table 5: Hallucination and relevance metrics for RaR-powered responses on the internal board-style dataset.** Summary of hallucination-related outcomes for the RaR method across all evaluated models on the internal unseen dataset (n = 65). "Context relevant" indicates the proportion of questions with clinically appropriate retrieved content. "Hallucination" refers to incorrect responses despite relevant context. "Correct despite irrelevant context" captures correct answers when the retrieved context was not useful. The final column reports the percentage of questions that were incorrect in zero-shot prompting but answered correctly with RaR.

| Model name | Context relevant | Hallucination (relevant context, incorrect response) | Correct despite irrelevant context | Zero-shot incorrect → RaR correct |
|---|---|---|---|---|
| Ministral-8B | 74% (48/65) | 6% (4/65) | 23% (15/65) | 29% (19/65) |
| Mistral Large (123B) | 74% (48/65) | 3% (2/65) | 25% (16/65) | 3% (2/65) |
| Llama3.3-8B | 74% (48/65) | 5% (3/65) | 20% (13/65) | 14% (9/65) |
| Llama3.3-70B | 74% (48/65) | 8% (5/65) | 25% (16/65) | 9% (6/65) |
| Llama3-Med42-8B | 74% (48/65) | 15% (10/65) | 14% (9/65) | 18% (12/65) |
| Llama3-Med42-70B | 74% (48/65) | 11% (7/65) | 17% (11/65) | 14% (9/65) |
| Llama4 Scout 16E | 74% (48/65) | 9% (6/65) | 26% (17/65) | 5% (3/65) |
| DeepSeek R1-70B | 74% (48/65) | 9% (6/65) | 26% (17/65) | 2% (1/65) |
| DeepSeek R1 (671B) | 74% (48/65) | 8% (5/65) | 25% (16/65) | 0% (0/65) |
| DeepSeek-V3 (671B) | 74% (48/65) | 5% (3/65) | 25% (16/65) | 2% (1/65) |
| Qwen 2.5-0.5B | 74% (48/65) | 32% (21/65) | 17% (11/65) | 29% (19/65) |
| Qwen 2.5-3B | 74% (48/65) | 9% (6/65) | 22% (14/65) | 12% (8/65) |
| Qwen 2.5-7B | 74% (48/65) | 8% (5/65) | 23% (15/65) | 17% (11/65) |
| Qwen 2.5-14B | 74% (48/65) | 8% (5/65) | 25% (16/65) | 11% (7/65) |
| Qwen 2.5-70B | 74% (48/65) | 5% (3/65) | 25% (16/65) | 3% (2/65) |
| Qwen 3-8B | 74% (48/65) | 11% (7/65) | 26% (17/65) | 5% (3/65) |
| Qwen 3-235B | 74% (48/65) | 9% (6/65) | 25% (16/65) | 2% (1/65) |
| GPT-3.5-turbo | 74% (48/65) | 8% (5/65) | 25% (16/65) | 22% (14/65) |
| GPT-4-turbo | 74% (48/65) | 9% (6/65) | 25% (16/65) | 15% (10/65) |
| o3 | 74% (48/65) | 9% (6/65) | 26% (17/65) | 9% (6/65) |
| GPT-5 | 74% (48/65) | 12% (8/65) | 23% (15/65) | 5% (3/65) |
| MedGemma-4B-it | 74% (48/65) | 9% (6/65) | 25% (16/65) | 17% (11/65) |
| MedGemma-27B-text-it | 74% (48/65) | 9% (6/65) | 25% (16/65) | 3% (2/65) |
| Gemma-3-4B-it | 74% (48/65) | 11% (7/65) | 25% (16/65) | 34% (22/65) |
| Gemma-3-27B-it | 74% (48/65) | 3% (2/65) | 25% (16/65) | 15% (10/65) |
| *Average* | *74% ± 0* | *9.2% ± 5.5%* | *23.5% ± 3.2%* | *11.8% ± 9.4%* |

**Supplementary Table 6: Response time comparison between zero-shot and RaR strategies on the internal dataset**. Average per-question response times (n=65) are reported in seconds as mean ± standard deviation for both individual models and aggregated model groups. On the internal dataset, a fixed overhead of 5,754.9 seconds per model, corresponding to context generation, was evenly distributed across all questions, contributing approximately 88.5 seconds per question. For time analysis, models were grouped based on parameter scale and architectural characteristics into six categories: the DeepSeek mixture of experts (MoE) group, the large model group (120–250B), the medium-scale group (~70B), the Gemma27B group, the small model group (7–8B), and the mini model group (3–4B). "Absolute difference" denotes the increase in average response time per question introduced by the RaR method, and "Relative increase" refers to the ratio of mean RaR time to mean zero-shot time per group. Final statistics are computed at the group level.

| Model / group name | Time | | | |
|---|---|---|---|---|
| | Zero-shot (s) | RaR (s) | Absolute difference (s) | Relative increase (times) |
| **DeepSeek-V3 group** | **65.0 ± 0.0** | **253.5 ± 0.0** | **188.5 ± 0.0** | **3.9 x** |
| **Large (120 – 250B) group** | **36.9 ± 16.8** | **216.7 ± 73.0** | **179.8 ± 72.3** | **5.9 x** |
| Llama4 Scout 16E | 36.3 ± 20.1 | 133.2 ± 20.4 | 96.8 ± 20.0 | 3.7 x |
| Mistral Large | 20.3 ± 10.1 | 249.1 ± 78.9 | 228.8 ± 71.2 | 12.3 x |
| Qwen 3-235B | 54.0 ± 28.7 | 267.8 ± 89.7 | 213.9 ± 79.2 | 5.0 x |
| **Medium (≈ 70B) group** | **36.5 ± 6.8** | **163.2 ± 22.7** | **126.6 ± 26.2** | **4.5 x** |
| DeepSeek R1-70B | 41.8 ± 23.7 | 173.1 ± 45.6 | 131.2 ± 41.4 | 4.1 x |
| Llama3-Med42-70B | 36.8 ± 18.1 | 133.2 ± 21.6 | 96.5 ± 20.8 | 3.6 x |
| Llama3.3-70B | 40.6 ± 20.7 | 160.0 ± 34.8 | 119.4 ± 31.3 | 3.9 x |
| Qwen 2.5-70B | 26.9 ± 14.9 | 186.4 ± 39.7 | 159.4 ± 35.3 | 6.9 x |
| **Gemma 27B group** | **53.7 ± 36.9** | **161.1 ± 54.3** | **107.4 ± 17.4** | **3.0 x** |
| Gemma-3-27B-it | 27.6 ± 13.2 | 122.7 ± 17.0 | 95.1 ± 16.0 | 4.4 x |
| MedGemma-27B-text-it | 79.8 ± 41.6 | 199.5 ± 53.3 | 119.7 ± 49.8 | 2.5 x |
| **Small (7 – 8B) group** | **10.3 ± 15.3** | **104.9 ± 11.0** | **94.6 ± 6.9** | **10.2x** |
| Llama3-Med42-8B | 2.4 ± 1.1 | 94.1 ± 2.5 | 91.7 ± 2.1 | 38.5 x |
| Llama3.3-8B | 5.9 ± 3.1 | 99.8 ± 5.5 | 93.8 ± 4.9 | 16.8 x |
| Ministral-8B | 2.9 ± 1.2 | 100.9 ± 5.8 | 98.0 ± 5.3 | 34.4x |
| Qwen 2.5-7B | 2.9 ± 1.3 | 106.8 ± 4.6 | 104.0 ± 4.0 | 37.2 x |
| Qwen 3-8B | 37.5 ± 20.8 | 123.0 ± 20.7 | 85.5 ± 20.7 | 3.3 x |
| **Mini (3 – 4B) group** | **7.7 ± 3.8** | **105.3 ± 6.5** | **97.6 ± 9.1** | **13.7 x** |
| Gemma-3-4B-it | 12.0 ± 5.0 | 100.2 ± 5.7 | 88.1 ± 5.6 | 8.3 x |
| MedGemma-4B-it | 6.3 ± 3.6 | 112.6 ± 14.5 | 106.3 ± 15.7 | 18.0 x |
| Qwen 2.5-3B | 4.8 ± 2.3 | 103.0 ± 3.8 | 98.2 ± 3.3 | 21.4 x |
| *Average* | *35.0 ± 22.9* | *167.5 ± 59.4* | *132.4 ± 41.7* | *6.9 ± 4.2 x* |

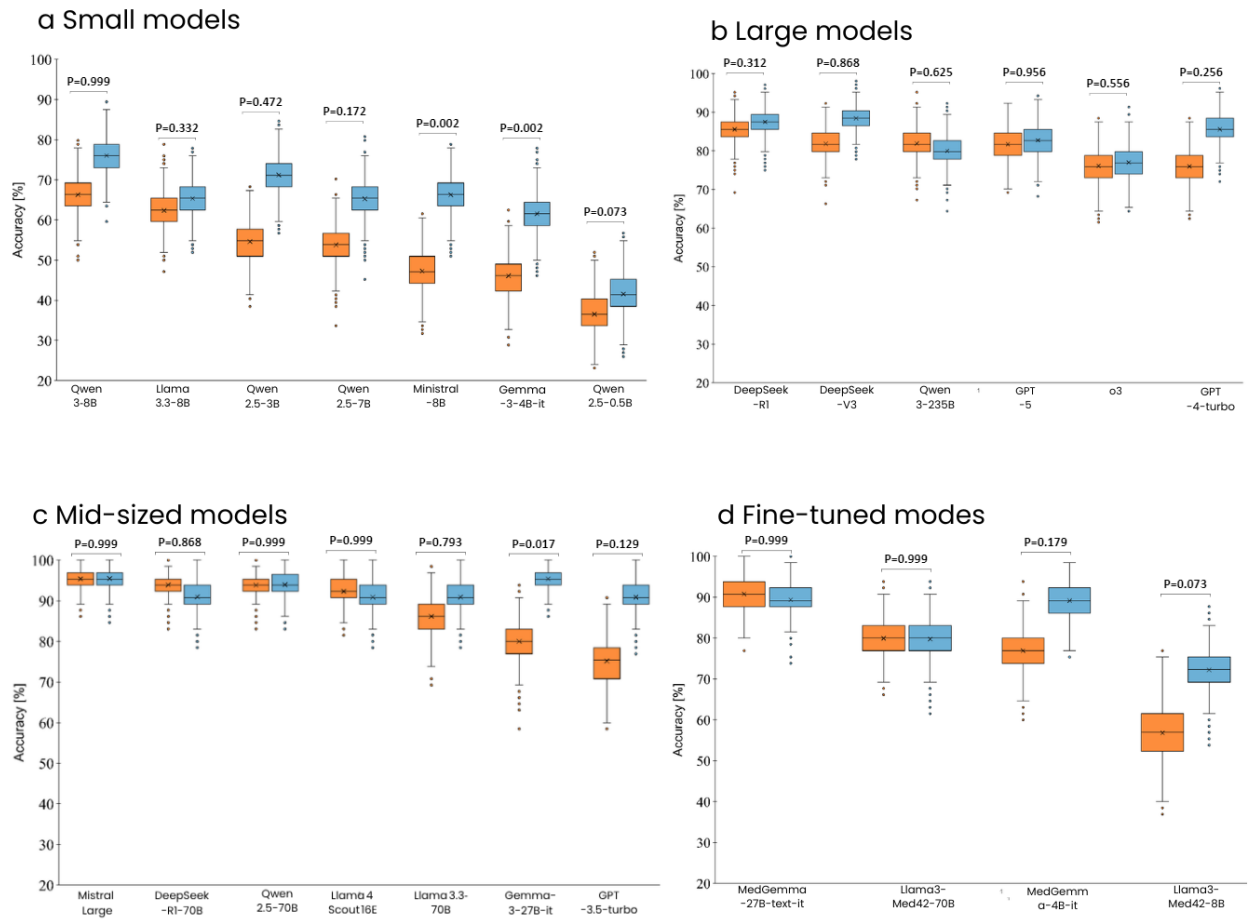

**Supplementary Figure 1: Comparative accuracy distributions for zero-shot versus RaR strategies across model groups on the internal dataset**. Accuracy results are shown for **(a)** small-scale models (Ministral-8B, Gemma-3-4B-it, Qwen 2.5-7B, Qwen 2.5-3B, Qwen 2.5-0.5B, Qwen 3-8B, Llama 3-8B), **(b)** large models (o3, GPT-5, DeepSeek-R1, Qwen 3-235B, GPT-4-turbo, DeepSeek-V3), **(c)** mid-sized models (Mid-Sized Models: GPT-3.5-turbo, Llama 3.3-70B, Mistral Large, Qwen 2.5-70B, Llama 4 Scout 16E, Gemma-3-27B-it, DeepSeek-R1-70B), **(d)** and medically fine-tuned models (MedGemma 27B-text-it, MedGemma 4B-it, Llama3-Med42-70B, Llama3-Med42-8B). comparisons were performed on the internal benchmark dataset (n =65). Boxplots display accuracy (%) distributions (n = 1 000) for zero-shot (orange) and RaR (blue): boxes span Q1–Q3, central line is the median (Q2), whiskers extend to 1.5×IQR and dots mark outliers. P-values were calculated between each pair's accuracy values for each model using McNemar's test on paired outcomes relative to RaR and adjusted for multiple comparisons using the false discovery rate. A p-value < 0.05 was considered statistically significant.

# Supplementary Note 1

*Diagnostic abstraction and retrieval preprocessing*

To enable structured, multi-step reasoning within the RaR (Retrieval and Reasoning) framework, we implemented a preprocessing step focused on diagnostic abstraction. This step ensured that retrieval operated on concise, clinically meaningful summaries of each question rather than on the full text, thereby improving retrieval precision and interpretability.

For every question in the RadioRAG dataset, we generated a short, comma-separated summary of its key diagnostic elements, typically including patient demographics, major imaging findings, clinical symptoms, and distinguishing modifiers (e.g., lesion type, location, or contrast pattern). These structured summaries served as retrieval queries to identify the most relevant clinical reports from Radiopaedia. To produce these abstractions, we used the Mistral Large (123B) model after comparing it against GPT-4o-mini and LLaMA-2-70B. Alternative models often produced verbose, incomplete, or redundant summaries, whereas Mistral Large consistently yielded concise, clinically faithful concept lists with strong semantic alignment to the original diagnostic intent. Its balance of brevity and coverage made it particularly effective for retrieval guidance within RaR.

**Representative cases**

### Case 1
**Question:** "A 45-year-old female with active HIV infection presented for a routine screening mammogram, which showed enlarged axillary lymph nodes but no other abnormalities. What is the most likely diagnosis?"
**Keywords generated by Mistral Large:** 45-year-old female, active HIV infection, routine screening mammogram, enlarged axillary lymph nodes, no other abnormalities.

### Case 2
**Question:** "A 15-year-old boy with scoliosis and recurrent nosebleeds has an incidental right middle lobe nodule."
**Keywords generated by Mistral Large:** 15-year-old male, no significant past medical history, scoliosis, frequent nosebleeds, dextroconvex thoracic scoliosis, incidental right middle lobe nodule.

### Case 3
**Question:** "A 61-year-old man with a painless, exophytic chest-wall mass of 18 months' duration that bleeds occasionally and shows spindle cells in a storiform pattern on biopsy."
**Keywords generated by Mistral Large:** 61-year-old man, painless fungating chest wall mass, 18 months' duration, occasional bleeding, vision impairment, spindle cells in storiform pattern, superficial exophytic soft tissue mass, no deep invasion.

**Case 4**
**Question:** "Strongly hyperdense lesions are seen in pulmonary arteries after vertebroplasty. What is the most likely diagnosis?"
**Keywords generated by Mistral Large:** hyperdense lesions in segmental pulmonary arteries of right lower lobe on non-contrast CT, history of vertebroplasty.

Diagnostic abstraction improved retrieval precision by reducing irrelevant matches caused by verbose or narrative phrasing. When applied across the full RadioRAG dataset, the retrieval pipeline incorporating these abstractions achieved clinically relevant context in 46% (48/104) of questions, as reported in the main text (**Table 3**). Although direct pre-abstraction retrieval rates were not logged, qualitative inspection confirmed that keyword-based queries produced more focused and semantically appropriate Radiopaedia matches, particularly for long, multi-clause questions. Overall, diagnostic abstraction provided a structured interface between natural-language questions and evidence retrieval, translating complex diagnostic phrasing into standardized clinical concept lists and enabling RaR to achieve more precise, interpretable, and contextually grounded reasoning.

# Supplementary Note 2

*Case studies: correct answers despite irrelevant or noisy retrieved context*

For each question in the RadioRAG dataset, a board-certified radiologist labeled the retrieved context as either clinically relevant or irrelevant, with the same label applied across all models for that item. Representative cases from the RaR runs of DeepSeek-R1 and GPT-3.5-turbo illustrate three characteristic reasoning patterns: context-independent correctness, where the model provides a correct answer despite irrelevant context; context-dependent grounded correctness, where the correct answer is supported by relevant evidence; and reasoning-shortcut errors, where the model gives an incorrect answer despite access to correct contextual information. These examples correspond to real RadioRAG items and are identified by their respective question IDs.

**Case 1 – Context-independent correctness**
**Question:** "Strongly hyperdense lesions are noted in the segmental pulmonary arteries of the right lower lobe on a non-contrast CT of the thorax. The patient has a history of vertebroplasty. What kind of lesions could that be?"
**Reference answer:** Pulmonary cement embolism
**Retrieved context:** Irrelevant
**Model outcomes (RaR):**
- DeepSeek-R1 — Correct
- GPT-3.5-turbo — Correct

Both models inferred the correct diagnosis from internal reasoning ("hyperdense pulmonary artery lesion + vertebroplasty → cement embolism"), without relying on the noisy retrieval. This exemplifies context-independent correctness, where accuracy arises from internal knowledge rather than the retrieved evidence.

**Case 2 – Grounded correctness**
**Question:** "A 61-year-old man presented with a painless, large, fungating chest wall mass of 18 months' duration that occasionally bled and impaired his vision of his feet. There were no systemic symptoms and staging CT of the chest, abdomen, and pelvis showed no other lesions. Biopsy revealed spindle cells in a storiform pattern infiltrating fat. MRI confirmed a superficial exophytic soft tissue mass confined to cutaneous and subcutaneous planes without deep invasion. What is the most likely diagnosis?"
**Reference answer:** Dermatofibrosarcoma protuberans
**Retrieved context:** Relevant
**Model outcomes (RaR):**
- DeepSeek-R1 — Correct
- GPT-3.5-turbo — Correct

This case shows an ideal evidence-grounded success. Both models effectively used the retrieved pathology and imaging cues to confirm the correct diagnosis.

**Case 3 – Reasoning shortcut error**
**Question:** "A 15-year-old male with no significant past medical history presents to orthopedic clinic for evaluation of scoliosis. An abnormality is seen on scoliosis films. Review of systems is positive for frequent nosebleeds. Frontal and lateral views of the spine show dextroconvex thoracic scoliosis with an incidental nodule in the right middle lobe. What is the most likely diagnosis?"
**Reference answer:** Pulmonary varix
**Retrieved context:** Relevant
**Model outcomes (RaR):**
- DeepSeek-R1 — Incorrect (predicted Pulmonary AVM)
- GPT-3.5-turbo — Incorrect

Both models defaulted to the familiar association "epistaxis → AVM," ignoring the key exclusion in the evidence. This represents a reasoning-shortcut error: failure to integrate a disconfirming retrieved fact.

These representative examples illustrate distinct patterns of reasoning under retrieval-augmented conditions. Some models can produce accurate answers even when retrieval is clinically irrelevant (Case 1), demonstrating robustness of internal medical knowledge. When evidence is relevant, RaR supports properly grounded reasoning by enabling structured integration of the retrieved content (Case 2). However, overreliance on prior associations can still lead models to disregard disconfirming evidence, resulting in errors despite correct retrieval (Case 3). Together, these cases show that RaR improves evidence use overall, yet retrieval relevance alone does not guarantee reasoning correctness, a key motivation for explicit grounding and auditing in clinical LLM pipelines.

# Supplementary Note 3

*Error analysis*

To better understand the reasoning patterns underlying model performance, we performed a qualitative error analysis across representative cases in the RadioRAG benchmark. Each question–answer pair was manually reviewed by a board-certified radiologist (TTN) to identify the dominant reasoning failure or success type. Errors were categorized into three principal types: reasoning shortcut errors, context integration errors, and context independence errors.

Reasoning shortcut errors occurred when models defaulted to familiar diagnostic associations instead of verifying all relevant details in the retrieved evidence.
- Example: In RSNA-RadioQA-Q59, all 25 models misdiagnosed a 15-year-old with frequent nosebleeds and a small lung nodule as having a pulmonary arteriovenous malformation. The report clearly described a pulmonary varix lacking an artery-to-vein connection, which should have excluded the diagnosis. Models relied on pattern familiarity ("nosebleeds → AVM") rather than applying the exclusion rule provided in the evidence. The radiologist made the same initial error without retrieval but corrected it after reviewing the RaR-retrieved report, confirming that the issue was reasoning, not data availability.

Context integration errors reflected failures to synthesize multiple correct elements into a single, coherent diagnosis.
- Example: In RSNA-RadioQA-Q5, which described a two-year-old with a large lung mass, most models focused narrowly on biopsy findings suggestive of a muscle-type tumor and ignored age and imaging clues indicating pleuropulmonary blastoma. The human expert correctly integrated these details to reach the right answer, demonstrating that the provided evidence was sufficient.

Context independence errors involved correct answers derived without meaningful use of the retrieved evidence.
- Example: In RSNA-RadioQA-Q88, the retrieved report discussed right-heart strain but also contained unrelated cardiac descriptions. Despite this, most models and the radiologist correctly diagnosed pulmonary embolism based on imaging clues such as right-ventricular enlargement, septal bowing, and contrast reflux. Likewise, in RSNA-RadioQA-Q97, all models correctly identified pulmonary cement embolism even though the retrieved report was classified as irrelevant. In both cases, models solved the question using internal medical knowledge rather than integrating the retrieved text, analogous to answering an open-book test without consulting the book.

Together, these analyses reveal that RaR mitigates reasoning shortcut and integration errors by enforcing structured, evidence-based reasoning, but context independence persists when retrieval adds noise rather than clarity. These findings highlight the need for future methods that not only improve factual accuracy but also ensure faithful use of supporting evidence. Overall, RaR contributes to more transparent and evidence-grounded diagnostic reasoning, even when retrieval context is imperfect.

# Supplementary Note 4

*Sensitivity/precision analysis and subgroup comparisons*

To contextualize non-significant findings and quantify precision, we performed subgroup-level paired analyses across models. For each dataset (RadioRAG, n = 104; internal n = 65), models were grouped as small, mid-sized, large, and clinically fine-tuned, consistent with the main manuscript.

The small subgroup included Ministral-8B, Gemma-3-4B-it, Qwen 2.5-7B, Qwen 2.5-3B, Qwen 2.5-0.5B, Qwen 3-8B, and Llama-3-8B. Mid-sized models comprised GPT-3.5-turbo, Llama 3.3-70B, Mistral Large, Qwen 2.5-70B, Llama 4 Scout 16E, Gemma-3-27B-it, and DeepSeek-R1-70B. Large-scale models included DeepSeek-R1, DeepSeek-V3, o3, Qwen 3-235B, GPT-4-turbo, and GPT-5. Clinically fine-tuned models consisted of MedGemma-27B-text-it, MedGemma-4B-it, Llama3-Med42-70B, and Llama3-Med42-8B. Qwen 2.5-14B was excluded as it did not align clearly with any predefined category.

Within each subgroup, mean accuracies under zero-shot and RaR conditions were computed, and paired differences were analyzed across models using two-sided paired t-tests. Reported statistics include the mean difference, its 95 % CI (t-based), p-value, and Cohen's dz.

On the RadioRAG dataset, RaR improved mean accuracy most notably for small models (+11.43 percentage points (pp), p = 0.002) and mid-sized models (+7.57 pp, p = 0.001), with a smaller and statistically non-significant effect in large models (+3.00 pp, p = 0.147). Clinically fine-tuned models also showed a consistent and significant gain (+8.75 pp, p = 0.001). On the internal generalization dataset, RaR produced a large and significant improvement in small models (+14.71 pp, p = 0.010), with positive but non-significant trends for mid-sized (+4.57 pp, p = 0.174) and clinically fine-tuned (+6.25 pp, p = 0.238) subgroups, and no measurable difference in large models (−0.17 pp, p = 0.953).

Overall, these subgroup analyses indicate that RaR's performance gains are most pronounced and statistically robust among smaller and mid-sized models, consistent with the main-text results. The lack of significance in other groups likely reflects limited sample size rather than the absence of true effect.